\documentclass[10pt,journal,compsoc]{IEEEtran}
\usepackage[utf8]{inputenc} 
\usepackage[T1]{fontenc}  

\usepackage{colortbl} 
\usepackage{xcolor}
\usepackage{color}

\usepackage{amsmath}
\usepackage{amssymb}
\usepackage[breaklinks=true,
            colorlinks,
            linkcolor = red,
            urlcolor  = purple, 
            citecolor = blue,
            bookmarks = black]{hyperref}  
\usepackage{url}  
\usepackage{amssymb}   
\usepackage{booktabs} 
\usepackage{amsfonts}  
\usepackage{nicefrac}  
\usepackage{microtype}  
\usepackage{wrapfig}
\usepackage{bbding}
\usepackage{appendix}
\usepackage{graphicx}
\usepackage{amsmath}
\usepackage{amsthm}
\usepackage{subcaption}
\usepackage{float}
\usepackage{appendix}
\usepackage{xcolor}
\usepackage{float} 
\usepackage{booktabs} %
\usepackage{multirow}
\usepackage{bm}
\usepackage{bbm}
\usepackage{makecell}
\usepackage{mathtools}
\usepackage[boxed,ruled]{algorithm2e}
\usepackage{amsmath}

\usepackage{xcolor}
\usepackage{tcolorbox}

\SetArgSty{textrm}
\ifCLASSOPTIONcompsoc
  \usepackage[nocompress]{cite}
\else
  \usepackage{cite}
\fi

\ifCLASSINFOpdf
\else
\fi

\begin{document}

%
\title{SurgLLM: A Versatile Large Multimodal Model with Spatial Focus and Temporal Awareness for Surgical Video Understanding}
%
%
%
%

\author{Zhen Chen$^*$, Xingjian Luo$^*$, Kun Yuan, Jinlin Wu$^\dagger$, Danny T.\,M. Chan, Nassir Navab,
\IEEEmembership{Fellow, IEEE}, \\Hongbin Liu, Zhen Lei, \IEEEmembership{Fellow, IEEE}, and Jiebo Luo \IEEEmembership{Fellow, IEEE}
\thanks{$^\dagger$\quad \ Corresponding author.}
\thanks{$^*$\quad \ Equal contribution.}
\IEEEcompsocitemizethanks{
\IEEEcompsocthanksitem Zhen Chen, Xingjian Luo, Jinlin Wu, Hongbin Liu, Zhen Lei, and Jiebo Luo are with Hong Kong Institute of Science \& Innovation, Hong Kong SAR, China (e-mail: zchen.francis@gmail.com; jinlin.wu@cair-cas.org.hk; hongbin.liu@cair-cas.org.hk; jluo@hkisi.org.hk).

\IEEEcompsocthanksitem Kun Yuan and Nassir Navab are with CAMP, Technische Universität München, Munich, Germany (e-mail: nassir.navab@tum.de).

\IEEEcompsocthanksitem Danny T.\,M. Chan is with the Department of Surgery, Faculty of Medicine, The Chinese University of Hong Kong, Hong Kong SAR, China (e-mail: tmdanny@surgery.cuhk.edu.hk).


}

}

\markboth{A Versatile Large Multimodal Model with Spatial Focus and Temporal Awareness for Surgical Video Understanding}%
{A Versatile Large Multimodal Model with Spatial Focus and Temporal Awareness for Surgical Video Understanding}

\IEEEtitleabstractindextext{%
\begin{abstract}
Surgical video understanding is crucial for facilitating Computer-Assisted Surgery (CAS) systems. Despite significant progress in existing studies, two major limitations persist, including inadequate visual content perception and insufficient temporal awareness in surgical videos, and hinder the development of versatile CAS solutions. In this work, we propose the SurgLLM framework, an effective large multimodal model tailored for versatile surgical video understanding tasks with enhanced spatial focus and temporal awareness. Specifically, to empower the spatial focus of surgical videos, we first devise Surgical Context-aware Multimodal Pretraining (Surg-Pretrain) for the video encoder of SurgLLM, by performing instrument-centric Masked Video Reconstruction (MV-Recon) and subsequent multimodal alignment. To incorporate surgical temporal knowledge into SurgLLM, we further propose Temporal-aware Multimodal Tuning (TM-Tuning)  to enhance temporal reasoning with interleaved multimodal embeddings. Moreover, to accommodate various understanding tasks of surgical videos without conflicts, we devise a Surgical Task Dynamic Ensemble to efficiently triage a query with optimal learnable parameters in our SurgLLM. Extensive experiments performed on diverse surgical video understanding tasks, including captioning, general VQA, and temporal VQA, demonstrate significant improvements over the state-of-the-art approaches, validating the effectiveness of our SurgLLM in versatile surgical video understanding. The source code is available at \url{https://github.com/franciszchen/SurgLLM}.
\end{abstract}
}

\maketitle

\IEEEdisplaynontitleabstractindextext

\IEEEpeerreviewmaketitle

\section{Introduction}\label{sec:introduction} 
\IEEEPARstart{S}{urgery} is at the core of modern healthcare systems, directly impacting patient outcomes and safety \cite{maier2017surgical}. Computer-Assisted Surgery (CAS) has emerged as a vital technology, augmenting surgeons with intraoperative guidance and analytical capabilities to enhance precision and mitigate risks \cite{chadebecq2023artificial, fiorini2022concepts}. In minimally invasive surgeries, surgeons rely heavily on endoscopic video feeds to perceive the surgical state and perform intricate actions \cite{haidegger2022robot}. These surgical videos, as visual records of surgical procedures, encode rich spatio-temporal and semantic information about instrument usage, tissue interactions, surgical workflow, decision making, and more. Consequently, developing multifaceted CAS technologies to thoroughly analyze surgical videos holds profound significance in improving the quality of surgery \cite{sznitman2012unified,giannarou2012probabilistic,tukra2021see}.

Recent efforts in CAS video analysis have made strides from various perspectives, including surgical scene understanding via anatomy segmentation and instrument detection \cite{jieboluo2024surgicalsam,chen2024asi}, modeling of procedural workflow through surgical phase recognition \cite{twinanda2016endonet,jin2021temporal,luo2023surgplan}, objective skill assessment by analyzing spatio-temporal patterns \cite{DBLP:journals/mva/ZhuLSK15,funke2019video,liu2021towards}, and knowledge extraction via automated operation narration \cite{xu2022rethinking,chen2023surgical} and visual question answering \cite{seenivasan2023surgicalgpt,bai2024surgical}. While these developments pave the way for impactful applications, they have primarily focused on developing specific algorithms or models for individual surgical tasks. This paradigm has resulted in a fragmented landscape of specialized tools, often lacking the flexibility to address the multifaceted nature of surgical procedures comprehensively. In summary, the CAS field has produced a collection of narrow, task-specific solutions rather than a unified approach for holistic analysis, limiting the potential for a versatile surgical video understanding system.

The emergence of multimodal large language models (MLLMs) \cite{llava,blip2,flamingo} offers a promising approach to addressing the limitations of current CAS video analysis. These models integrate the natural language capabilities of large language models (LLMs)\cite{llama2,vicuna,mixtral} and the visual perception of visual encoders via tailored multimodal connectors \cite{flamingo,bai2023qwen,dai2023instructblip,liu2024visual,cambrian,honeybee}. Pioneering MLLM studies have initially focused on diverse image-based tasks, demonstrating remarkable versatility across various domains. Building upon these image-based foundations, video-capable MLLMs such as VideoChat\cite{videochat}, VideoLLaMA\cite{videollama}, Qwen-VL\cite{Qwen2.5-VL}, InternVL\cite{internvl2.5}, and LLaMA-VID\cite{llamavid} have shown promising results in video comprehension tasks, including video captioning, visual question answering, and temporal action localization. These advancements reveal the potential for applications in complex visual scenarios. However, despite their successes in general video understanding, these MLLMs face significant challenges when applied directly to surgical video analysis due to the unique characteristics of minimally invasive surgeries.


The first challenge arises from the existing representation learning paradigms, leading to the inadequacy of visual content perception in surgical videos. Current MLLMs rely heavily on visual encoders pre-trained on natural scenarios \cite{videochat,videollama,Qwen2.5-VL,internvl2.5}, which struggle when applied directly to surgical videos due to the fundamental differences between surgical and natural scene videos. Surgical videos exhibit distinctive visual characteristics that general-purpose visual encoders struggle to capture effectively. On one hand, the action-centric video dynamics of surgical videos engender complex foreground-background relationships that existing visual pretraining techniques (\textit{e.g.}, multimodal contrastive learning \cite{clip} and conventional masking strategies \cite{videomae}) fail to capture adequately. These video dynamics are characterized by focused instrument movements against a relatively static background, a scenario rarely encountered in natural video datasets. On the other hand, the substantial visual redundancy in surgical videos is typified by long sequences of visually similar frames interspersed with critical operations or anomalous events that require rapid detection \cite{jin2021temporal,luo2024surgplan,chen2023temporal}. This presents a unique challenge in maintaining model attention over prolonged periods while simultaneously ensuring high responsiveness to sudden, significant changes. Therefore, two key improvements are required to adapt the existing MLLM architecture for surgical videos, including developing surgical-specific masking strategies to better capture foreground-background dynamic relationships and enhancing multi-scale embedding techniques to maintain high sensitivity to critical events in lengthy surgical videos.


The second challenge is caused by the insufficient temporal awareness capabilities of current MLLMs within the surgical context  \cite{videochat,videollama,Qwen2.5-VL,internvl2.5}. The clinical nature of surgery demands precise temporal awareness, a requirement that current video LLMs fail to meet adequately. In real-world surgical practice, precise timing is crucial for various applications, including efficient scheduling of senior surgeons and coordinating surgical team activities \cite{schmitt2022team,tiferes2016loud}. While existing video LLMs excel in general video understanding tasks \cite{Qwen2.5-VL,internvideo,internvl2.5}, they often lack the required fine-grained temporal awareness, especially when processing surgical videos. Specifically, existing video LLMs struggle to accurately associate surgical actions or events with exact timestamps, fail to fully comprehend the unique temporal dependencies in surgical procedures, and perform poorly in providing real-time insights or assisting with time-critical decision-making. These limitations significantly constrain the potential application of MLLMs in surgical environments, impeding their integration into clinical workflows. Therefore, the MLLMs for surgical videos are expected to possess enhanced temporal reasoning capabilities, including accurate identification and localization of critical time points during surgery, understanding of temporal relationships between different surgical stages, and real-time prediction of surgical progress for timely decision support.

To address these challenges and advance the field of surgical video analysis, we propose an effective framework named SurgLLM that tailors the large multimodal model for comprehensive surgical video understanding. To create a unified, versatile system capable of handling the multifaceted nature of surgical procedures, our SurgLLM is designed to overcome the limitations of current MLLMs when applied to the surgical domain. Specifically, our SurgLLM framework comprises three key innovations, including Surgical Context-aware Multimodal Pretraining (Surg-Pretrain), Temporal-aware Multimodal Tuning (TM-Tuning), and a Surgical Task Dynamic Ensemble. Specifically, Surg-Pretrain first addresses the challenge of inadequate visual content perception in surgical videos by introducing Multi-scale Instrument-centric Masked Video Reconstruction (MV-Recon) at varying temporal scales and the subsequent surgical multimodal alignment. Then, TM-Tuning tackles the issue of insufficient temporal awareness by implementing the textural-visual temporal interleave embeddings. Finally, the Surgical Task Dynamic Ensemble enables the model to efficiently handle diverse surgical tasks without compromising performance on individual subtasks.

The contributions of this work are summarized as follows:
\begin{itemize}
    \item We propose SurgLLM tailored for surgical video understanding, integrating spatial focus and temporal awareness to address the unique challenges of surgical scenes that general-purpose video LLMs fail to handle effectively.
    
    \item We propose Surg-Pretrain, consisting of MV-Recon that captures the unique foreground-background dynamics of surgical videos, combined with surgical video context alignment to enhance surgical scene understanding capabilities.
    
    \item We devise TM-Tuning that tightly couples temporal information with textual-visual temporal interleave embeddings, enabling precise temporal reasoning for surgical video understanding.

    \item We propose the Surgical Task Dynamic Ensemble to efficiently adapt to diverse surgical tasks, addressing the challenge of task diversity in surgical video analysis while preventing catastrophic forgetting.
    
    \item Extensive experiments demonstrate significant improvements over state-of-the-art methods across captioning, general VQA, and temporal VQA tasks, validating its potential as a versatile tool for computer-assisted surgery.
    
\end{itemize}

The rest of this paper is organized as follows. In Section~\ref{sec:related_work}, we review the literature related to this paper. In Section~\ref{sec:method}, we discuss the technical details of the proposed SurgLLM step by step. Extensive experiments and ablation studies are presented in Section~\ref{sec:experiments}. Finally, we conclude this paper in Section~\ref{sec:conclusion}.


\section{Related Work}\label{sec:related_work}
\subsection{Surgical Scene Understanding}
Surgical scene understanding has become a critical research area, encompassing a wide range of tasks to interpret the complex dynamics of surgical environments \cite{surgicalds,aiinsurgery,chen2024vs}. To enable models to better learn surgical characteristics and handle diverse scene understanding tasks, multiple annotated datasets have been developed in collaboration with professional surgeons. Notable examples include Cholec80 \cite{cholec80}, CholecT50 \cite{cholect50}, EndoVis2017 \cite{endovis2017}, and EndoVis2018 \cite{endovis2018}, which provide rich annotations to facilitate model training and evaluation. However, these datasets are predominantly labeled on a per-frame basis, lacking the question-answer pair annotations required for Multimodal Large Language Models (MLLMs). This limitation restricts their applicability to traditional tasks and hinders their use in more holistic, interactive applications. To address this limitation, VQA datasets like SurgicalVQA \cite{surgvqa} and SGG-VQA \cite{sggvqa} have been introduced. These datasets incorporate question-answer pairs for visual question answering (VQA) tasks. However, they remain confined to frame-level annotations and fail to provide video-level insights, such as capturing dynamic changes and temporal dependencies throughout surgical procedures. This gap highlights the need for datasets and methods that can facilitate a deeper understanding of the surgical scene beyond static frame analysis.

In parallel, recent advances in surgical scene understanding have largely focused on single-task challenges, such as surgical triplet detection \cite{triplet}, instrument segmentation and detection \cite{jieboluo2024surgicalsam,sznitman2012unified}, as well as surgical motion assessment~\cite{DBLP:journals/mva/ZhuLSK15}. Triplet detection, which involves identifying relationships between surgical instruments, anatomical structures, and actions, has been explored using graph-based methods and transformer architectures to capture complex spatial and temporal dependencies. Similarly, instrument segmentation and detection have achieved significant progress with the adoption of deep learning techniques, including region-based convolutional neural networks \cite{rcnn,fastrcnn,fasterrcnn,yolo} and ViT-based architectures \cite{vit}, enabling precise localization and delineation of surgical tools in both 2D and 3D spaces \cite{nnunet}. Despite these advancements, such single-functional tasks lack versatility and are not integrated into a unified system capable of addressing the multifaceted needs of a surgical environment.

Building on these efforts, surgical video understanding has emerged to incorporate temporal information for enhanced analysis of surgical workflows \cite{realtimesurgicalvideo,surgicalvideo}. For instance, phase recognition, a key task in this domain, leverages recurrent neural networks and temporal convolutional networks to model the sequential nature of surgical procedures, achieving promising results in workflow understanding \cite{luo2023surgplan}. Additionally, video-based instrument segmentation \cite{wu2024real} extends static segmentation techniques by considering temporal consistency, employing methods such as optical flow and spatiotemporal attention mechanisms. Surgical video captioning, another emerging area, uses encoder-decoder architectures with attention mechanisms to generate descriptive summaries of surgical actions and events \cite{chen2023surgical}. These video-based advancements collectively contribute to a more holistic understanding of surgical scenes, enabling applications in computer-assisted interventions and potentially improving surgical outcomes.

\subsection{Multimodal Large Language Models}
Multimodal Large Language Models (MLLMs) have emerged as a promising approach to address the limitations of current CAS video analysis approaches \cite{agisurgical}. In the domain of image understanding, several innovative models have demonstrated remarkable capabilities. LLaVA \cite{llava} utilizes linear connection layers and adheres to a pretraining and instruction fine-tuning paradigm. BLIP-2 \cite{blip2} introduces the Q-Former, an innovative mechanism to extract image information through learnable queries. InstructBLIP \cite{instructblip} further augments this by computing attention between the Q-Former and an instructor, facilitating more focused, instruction-relevant target identification. Additional noteworthy approaches include learnable query methods in QwenVL \cite{bai2023qwen},  interleaved image-text architectures in Flamingo \cite{flamingo}, and multi-scale feature extraction techniques in Cambrian-1 \cite{cambrian}. These models generally integrate pre-trained vision encoders (\textit{e.g.}, ViT \cite{vit}, CLIP \cite{clip}) with large language models (\textit{e.g.}, LLaMA-2 \cite{llama2}, Vicuna \cite{vicuna}) via diverse multimodal connectors, demonstrating the adaptability of MLLMs in addressing a wide range of image-based tasks.

Building upon these image-based foundations, video-capable MLLMs have made significant strides in addressing the temporal aspects of video comprehension\cite{videollmsurvey}. Models such as VideoChat \cite{li2023videochat} and ChatVideo \cite{wang2023chatvideo} leverage external models and databases to convert video and audio information into text, which is then processed by language models. VideoLLaMA \cite{videollama} employs a Q-Former to process features from each frame, combining them through linear layers. VideoLLaVA \cite{videollava} extends the LLaVA approach to video, using a LanguageBind\cite{languagebind} encoder followed by linear projection into the LLM. More recent advancements include VideoLLaMA v2 \cite{videollamav2}, which incorporates a Spatial-Temporal Convolution connector for better spatio-temporal perception, and VTimeLLM \cite{vtime}, which injects temporal awareness by prefixing frame information. TimeChat \cite{timechat} takes a unique approach by combining the query with time-stamped instructions before attention computation. ChatUni \cite{chatuni} represents videos as a set of dynamic visual tokens by a clustering algorithm. Despite these advancements, current MLLMs still face challenges in fully analyzing surgical videos due to the domain's unique attributes, including complex ego-centric views and the need for precise temporal awareness in clinical contexts. Our proposed SurgLLM framework addresses these limitations through targeted design choices, making it particularly well-suited for various surgical video analysis tasks.


\begin{figure*}[t]
    \centering
    \includegraphics[width=0.99\textwidth]{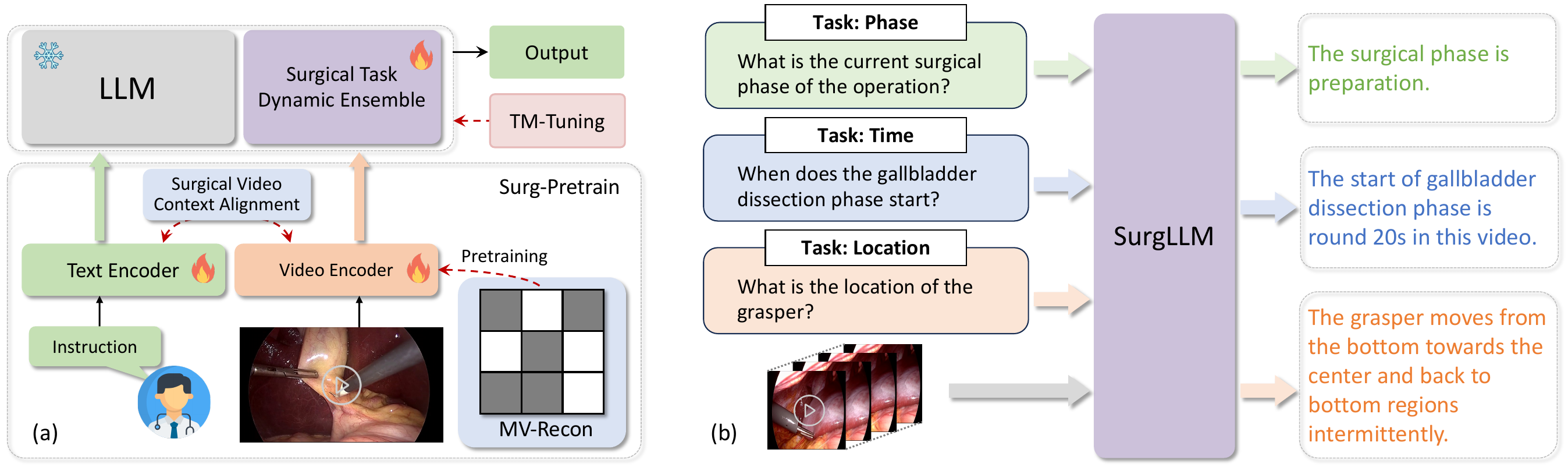}
    \caption{(a) The overview of the SurgLLM training pipeline, including Surgical Context-aware Multimodal Pretraining (Surg-Petrain) for video and text encoders and Temporal-aware Multimodal Tuning (TM-Tuning) for Surgical Task Dynamic Ensemble. (b) The overview of the SurgLLM inference pipeline. The well-trained SurgLLM adaptively utilizes the multi-task Q-Former and dynamic ensemble of task-specific LoRA weights for versatile surgical video understanding tasks, including phase recognition, temporal localization, and instrument analysis.}
    \label{fig1_framework}
\end{figure*}

\section{Methodology}\label{sec:method}
\subsection{Overview}

In this work, we present the SurgLLM framework for the comprehension of surgical videos, as illustrated in Fig. \ref{fig1_framework}. First, we propose a \textbf{Surgical Context-aware Multimodal Pretraining (Surg-Pretrain)} to integrate multi-scale surgical instrument perception capabilities and text alignment. Second, we introduce a \textbf{Temporal-aware Multimodal Tuning (TM-Tuning)} designed to enhance the MLLM's capacity to discern temporal information in videos. Finally, we propose a \textbf{Surgical Task Dynamic Ensemble} that empowers SurgLLM to more effectively address tasks demanding diverse aspects of capability. These methodologies are engineered to optimize the capacity of SurgLLM to interpret surgical video content and respond to corresponding queries with enhanced accuracy and depth.

\subsection{Surgical Context-aware Multimodal Pretraining}

To improve the video encoder in SurgLLM with surgical-specific visual perception, we propose a surgical context-aware multimodal pretraining (Surg-Pretrain) consisting of two steps: instrument-centric Masked Video Reconstruction (MV-Recon) and surgical video context alignment, as illustrated in Fig. \ref{fig2-surg-pretrain}. In the first step, we introduce a multi-scale instrument-centric tube masking strategy that prioritizes masking regions containing surgical instruments, and devise multi-scale tube masking across varying temporal durations to address visual redundancy in surgical videos. The second step bridges the learned visual representations with surgical textual knowledge through contrastive learning.

\subsubsection{Instrument-centric Masked Video Reconstruction}
To comprehend the surgical foreground and background, we first propose a multi-scale instrument-centric tube masking technique for the video encoder, thereby better capturing crucial dynamic information during surgery. Furthermore, to address the redundancy issue in surgical videos, we devise a multi-scale mask reconstruction, enabling the video encoder of our SurgLLM to undergo comprehensive pretraining across various temporal durations.

\begin{figure*}[t]
    \centering
    \includegraphics[width=0.97\textwidth]{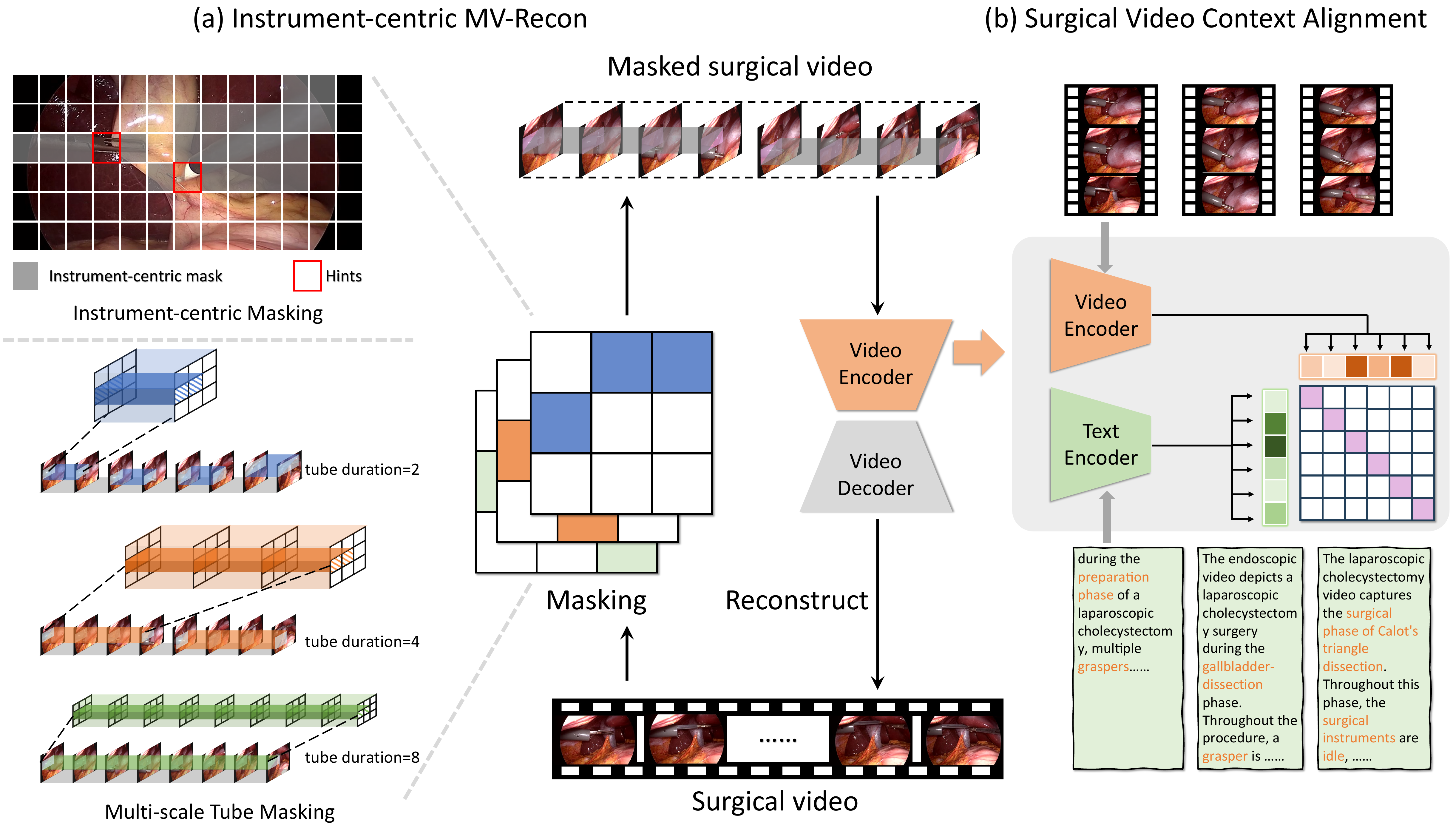}
    \caption{Illustration of Surg-Pretrain for SurgLLM. (a) Multi-scale Instrument-centric Masked Video Reconstruction adopts instrument-focused masking with varying temporal scales to capture surgical dynamics while addressing visual redundancy. (b) Surgical Video Context Alignment learns surgical-specific visual representations and aligns them with textual descriptions through contrastive learning, enabling SurgLLM to understand complex surgical scenes and instrument interactions.}
    \label{fig2-surg-pretrain}
\end{figure*}

\noindent \textbf{Multi-scale Instrument-centric Tube Masking}.
Unlike VideoMAE \cite{videomae} that employs random masking for natural videos, surgical videos exhibit distinct foreground-background separation where surgical instruments represent the most critical visual elements. Therefore, we propose an instrument-centric tube masking approach that prioritizes regions with ongoing procedures for our SurgLLM, as illustrated in Fig. \ref{fig2-surg-pretrain} (b).

Given a surgical video $\bm{v} \in \mathbb{R} ^{ N \times H\times W \times C}$ as input,  where $N$ is the number of frames, and $H, W, C$ are the height, width, and channel number. We first divide it into video tubes at multiple temporal scales to address the inherent visual redundancy in surgical procedures. Specifically, we generate each video tube $\bm{T} \in \mathbb{R} ^{k \times h \times w \times C}$ with varying temporal duration $k$, where $h$ and $w$ are the height and width of the video tube $\bm{T}$. This multiscale tube strategy enables the video encoder to perceive temporal features at different granularities, from fine-grained instrument movements to broader procedural patterns.

Then, we select the first frame of each tube as a reference, and set the instrument mask indicator $\bm{M}$ to indicate the masking for each tube with respect to the reference frame, as follows:
\begin{equation}
\bm{M}_{i} = \begin{cases}
1, & \text{if } \bm{T}_{i} \text{ contains instruments}, \\ 0,& \text{otherwise},
\end{cases}
\end{equation}
where $\bm{M}_{i}$ is the instrument mask indicator for the $i$-th tube $\bm{T}_{i}$. In this way, the instrument mask indicator $\bm{M}$ is 1 if the tube is involved with surgical instruments, otherwise it is 0.

Among the mask indicator $\bm{M}$ containing surgical instruments, we further randomly retain a small proportion $r$ of video tubes as hints for MV-Recon to reconstruct the surgical
video better, while masking the rest of the video tubes. Specifically, we randomly mark the video tubes with $\bm{M}$ as 1 with probability $r$ as a hint. In contrast, we mark the remaining video tubes as 0, erase them at the input, and use them as the target of reconstruction. As such, we define the hint indicator $\bm{H}$ as follows:
\begin{equation}
\bm{H}_{i} = \begin{cases}
1, & \text{if tube } \bm{T}_i  \text{ is selected as a hint and } \bm{M}_i=1, \\ 0, & \text{otherwise.}
\end{cases}
\end{equation}
In this way, we use $\bm{H}$ to represent the surgical video as input, and identify the visible video tubes with $\bm{H}=1$ that serve as hints for MV-Recon, while the remaining video tubes with $\bm{H}=0$ are the reconstruction targets.

\noindent \textbf{Masked Tube Reconstruction}.
On the basis of multi-scale instrument-centric tube masking, we further conduct the surgical video masked tube reconstruction with the autoencoder scheme. The autoencoder comprises a ViT-based video encoder $\mathcal{E}_{v}$ and a video decoder $\mathcal{D}_{v}$.
For the input surgical video $\bm{v}$, we first divide it into a set of 3D volumes $\mathcal{P}=\{p_j\}_{j=1}^m$. Based on the hint indicator $\bm{H}$, these volumes are partitioned into the visible volumes $\mathcal{P}_{\rm vis}$ and the masked volumes $\mathcal{P}_{\rm mask}$. If a 3D volume $p_j$ is located within the area where $H_j = 1$, it belongs to the visible volumes $\mathcal{P}{\rm vis}$, otherwise it is assigned to the masked volumes $\mathcal{P}_{\rm mask}$.

Then, the video encoder $\mathcal{E}_{v}$ processes only the visible volumes $\mathcal{P}_{\rm vis}$, and the video decoder $\mathcal{D}_{v}$ further reconstructs the masked volumes $\mathcal{P}_{\rm mask}$ from the encoded representation. The reconstruction process can be formulated as follows:
\begin{equation}
    \hat{\mathcal{P}}_{\rm mask} = \mathcal{D}_{v}(\mathcal{E}_{v}(\mathcal{P}_{\rm vis})),
\end{equation}
where $\hat{\mathcal{P}}_{mask}$ denotes the set of reconstructed volumes.

Finally, the volume-wise Mean Squared Error (MSE) loss is calculated between the original masked volumes and the reconstructed ones:
\begin{equation}
 \mathcal{L}_{\rm recon} = \text{MSE}(\mathcal{P}_{\rm mask}, \hat{\mathcal{P}}_{\rm mask}).
\end{equation}
In this way, this self-supervised pretraining process enables the video encoder $\mathcal{E}_{v}$ to produce high-quality visual features of surgical videos, emphasizing surgical instruments and exploiting surgical content in subsequent steps.

\subsubsection{Surgical Video Context Alignment}
To bridge the learned visual representations with surgical textual knowledge, we further perform the surgical video context alignment using multimodal contrastive learning techniques \cite{clip,li2023unmasked}. By leveraging the video encoder $\mathcal{E}_{v}$ pretrained from the instrument-centric masked video reconstruction, this alignment step aligns the visual features with surgical procedure descriptions, as shown in Fig. \ref{fig2-surg-pretrain} (b). Specifically, we employ three complementary objectives to achieve multimodal context alignment, including the video-to-text contrastive learning (VTC) that learns global video-text correspondences, video-to-text matching (VTM) that performs fine-grained similarity assessment, and masked language modeling (MLM) that enhances textual understanding within multimodal context.

Given a batch of $K$ video-text pairs, we first extract features using the pretrained video encoder $\mathcal{E}_{v}$ and a text encoder $\mathcal{E}_{t}$. These features are then projected into a shared embedding space via learnable projection layers $W_v$ and $W_t$. The final aligned and normalized embeddings are computed as follows:
\begin{equation}
\begin{split}
f_v &= \mathcal{N}(\mathcal{E}_{v}(V) \cdot W_v),\\
f_t &= \mathcal{N}(\mathcal{E}_{t}(D) \cdot W_t),
\end{split}
\end{equation}
where $V = \{v_1, v_2, ..., v_K\}$ represents the input surgical videos, and $D = \{d_1, d_2, ..., d_K\}$ denotes their corresponding dense procedural captions. The normalization function $\mathcal{N}$ ensures feature consistency across modalities. In this way, the surgical video context alignment enables SurgLLM to associate visual patterns with high-level surgical semantics, providing robust multimodal representations for downstream surgical reasoning tasks.

\begin{figure*}[t]
    \centering
    \includegraphics[width=0.97\textwidth]{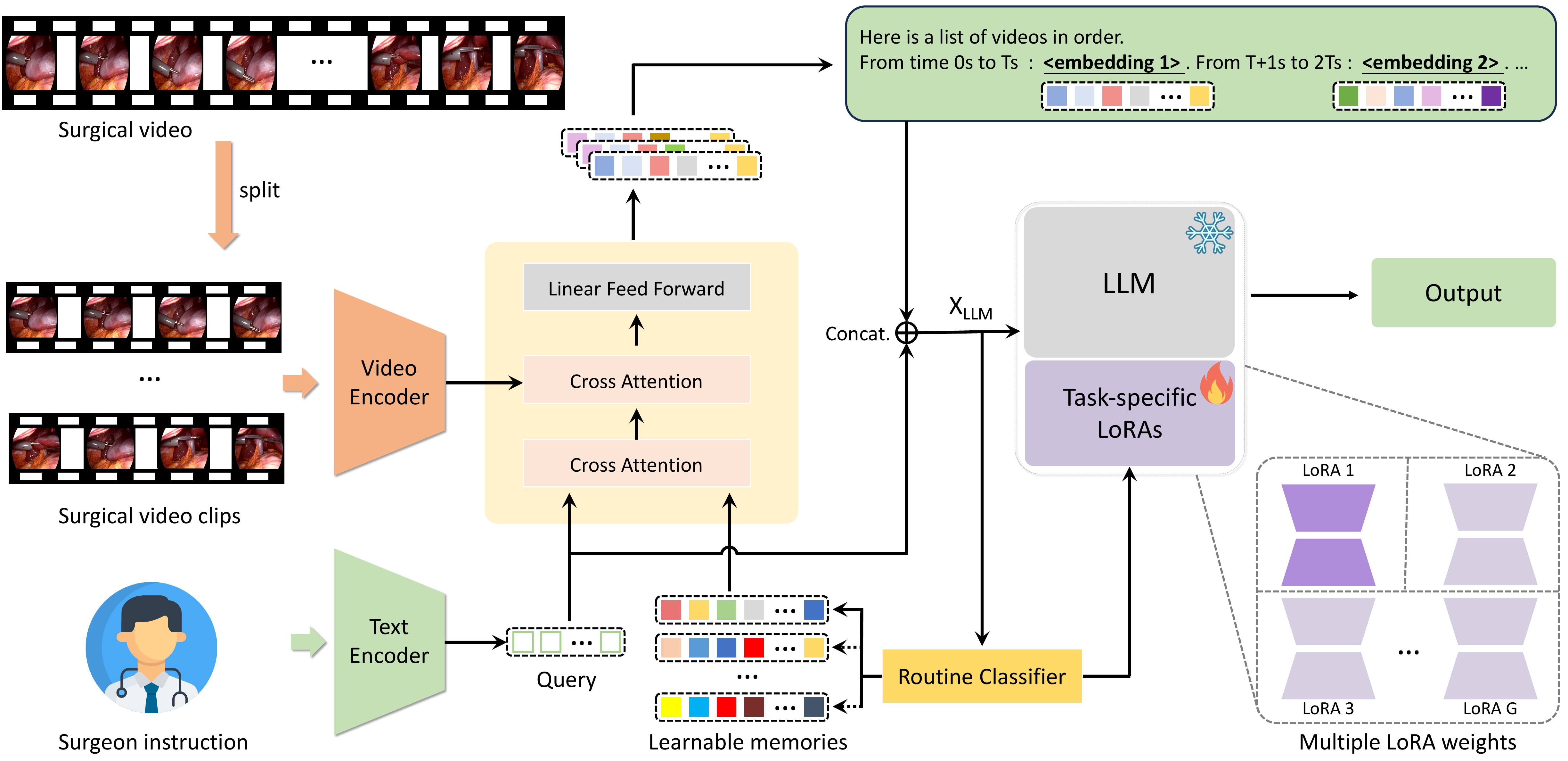}
    \caption{Illustration of Temporal-aware Multimodal Tuning (TM-Tuning) and the Surgical Task Dynamic Ensemble. (a) TM-Tuning splits the input video into temporal segments, processes them through the video encoder, and creates interleaved embeddings with temporal descriptors. (b) The Surgical Task Dynamic Ensemble adopts multiple task-specific learnable memories and corresponding LoRA weights selected by task routing, enabling SurgLLM to adaptively handle diverse surgical tasks, including phase recognition, location detection, and temporal analysis.}
    \label{fig3-tuning}
\end{figure*}

\begin{figure}[t]
    \centering
    \includegraphics[width=0.49\textwidth]{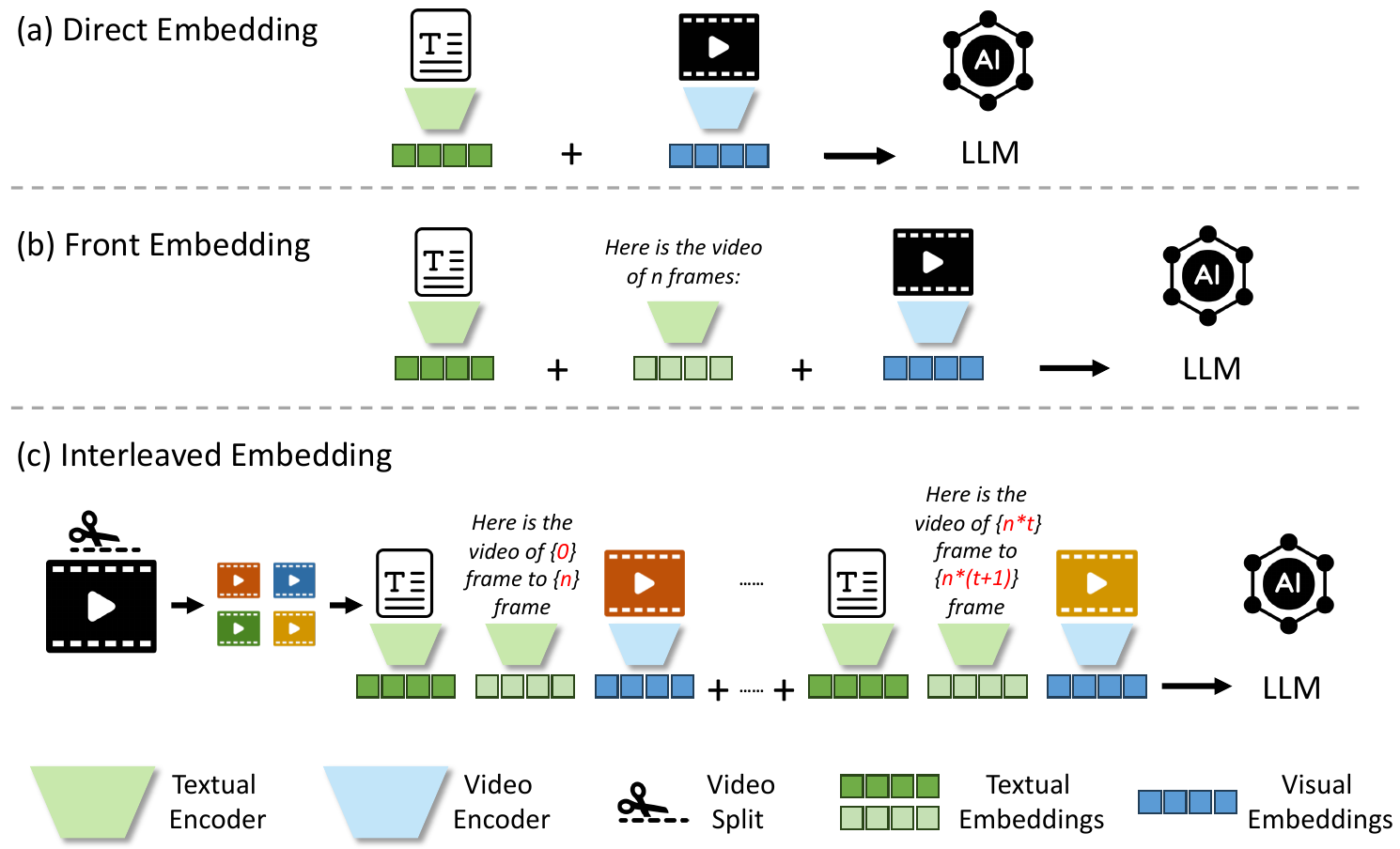}
    \caption{Comparison of temporal embedding strategies for surgical video understanding. (a) \textbf{Direct embedding}: The baseline that directly inputs both textual and visual tokens into the LLM without explicit connections between them. 
    (b) \textbf{Front embedding}: Add the text description of the visual content before visual tokens. 
    (c) \textbf{Interleaved embedding}: Divide the visual information into multiple segments, where each visual segment is preceded by its corresponding text description.}
    \label{fig4-embedding}
\end{figure}

\subsection{Temporal-aware Multimodal Tuning}

Surgical videos often span extended durations with complex temporal dynamics, posing significant challenges for multimodal LLMs in accurate temporal reasoning. Existing approaches, such as VTimeLLM \cite{vtime}, prepended the video duration (\textit{e.g.}, \textit{"This is a video with 100 frames"}) to the video embeddings, as shown in Fig.~\ref{fig4-embedding} (a) and (b). However, this leads to a substantial gap between temporal descriptions and corresponding visual features, particularly for temporally distant video segments, weakening the temporal perception due to long-distance attention dependencies.

To address this limitation, we propose the Temporal-aware Multimodal Tuning (TM-Tuning) by tightly coupling temporal information with visual features throughout the video sequence, as illustrated in Fig. \ref{fig3-tuning}. Specifically, we first segment the input surgical video $\bm{v}$ into $N$ sequential clips $\{\bm{c}_i\}_{i=1}^N$, as elaborated in Fig. \ref{fig4-embedding} (c). Then, these clips are processed through our pretrained video encoder and a visual adapter to obtain visual feature tokens $H_v^i$ for each temporal segment $i$. 
For each video clip $\bm{c}_i$, we generate corresponding temporal descriptors $S_i$ that explicitly encode its temporal boundaries as \textit{"This is a video clip spanning from $i \times t$ to $(i+1) \times t$ seconds"}, where $t$ denotes the clip duration. After that, these descriptors are interleaved with their corresponding visual features, formulating the final input sequence for the LLM as follows:
\begin{equation}\label{eq_llm_input}
X_{\rm LLM} = [S_{1}, H_v^1, S_{2}, H_v^2, \ldots, S_{N}, H_v^N, q],
\end{equation}
where $q$ is the query regarding this surgical video. As such, this interleaved structure ensures that each visual segment $H_v^i$ is immediately preceded by its temporal context $S_i$, enabling direct association between temporal attributes and visual content. By maintaining close proximity between temporal descriptors and their corresponding visual features, our temporal-aware multimodal instruction enhances the capability of SurgLLM to perform accurate temporal reasoning and respond to time-sensitive surgical queries.

\subsection{Surgical Task Dynamic Ensemble}

Surgical video analysis encompasses diverse tasks such as instrument recognition, phase classification, and procedural reasoning, each requiring specialized understanding. Traditional fine-tuning approaches face a fundamental challenge, \textit{i.e.}, optimizing for one task often degrades performance on others, resulting in mutual constraints that limit the overall effectiveness. To overcome this limitation, we propose the Surgical Task Dynamic Ensemble to dynamically adapt the components of our SurgLLM based on task requirements, as illustrated in Fig. \ref{fig3-tuning}.

Specifically, the Surgical Task Dynamic Ensemble adopts a multi-task Q-Former as the visual adapter, which contains multiple sets of task-specific learnable memories to bridge the latent space between the video encoder $\mathcal{E}_{v}$ and the LLM regarding different surgical tasks. Given a surgical task $g$, the multi-task Q-Former adaptively utilizes a specific learnable memory $Q_g\in \mathbb{R}^{C_{\rm embed}}$ to interact with the visual features $z_v$ from the video encoder. This process is formulated as:
\begin{equation}
    H_v^{g} = \text{Linear}(\phi(\phi(Q_g, q), z_v)),
\end{equation}
where $H_v^{g} \in \mathbb{R}^{C_{\rm embed}}$ is the processed visual tokens that integrate task-relevant visual information, $\phi$ is the cross attention calculation regarding the query $q$, and $\text{Linear}$ denotes the linear layer. With $G$ learnable memories $\{Q_g\}_{g=1}^{G}$, the output $H_v$ captures different aspects of the visual content, serving as the input for the LLM in Eq. \eqref{eq_llm_input}.

\begin{figure}[t]
    \centering
    \includegraphics[width=0.40\textwidth]{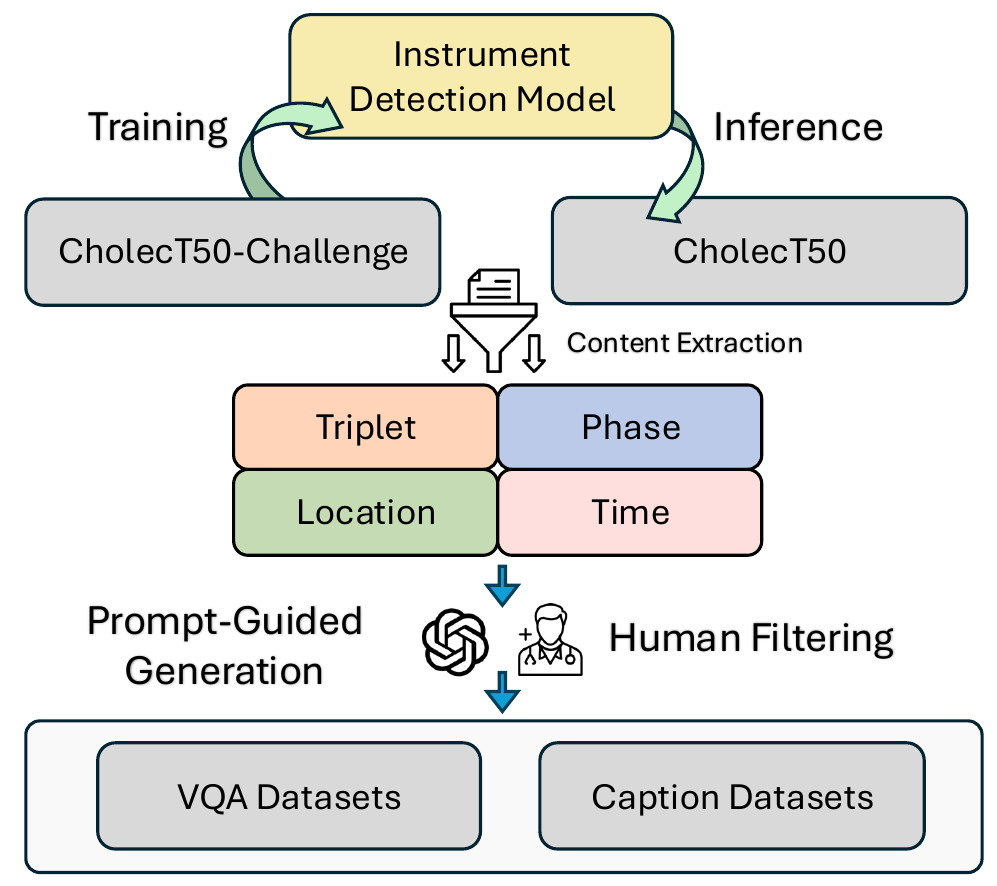}
    \caption{The dataset construction pipeline for comprehensive surgical video understanding. Starting from the CholecT50 annotations, we employ the instrument detection model trained on partial bounding box data to generate complete location annotations. The pipeline integrates triplet content extraction with GPT-4-guided VQA generation, followed by human filtering to create high-quality caption datasets that cover diverse information, including the triplet, phase, location, and time, for versatile surgical video analysis.}
    \label{fig5}
\end{figure}

Furthermore, to enable dynamic adaptation on diverse tasks, we utilize a lightweight classifier $\mathcal{C}$ that categorizes the query $q$ within interleaved embeddings $X_{\rm LLM}$ into one of the task routines $g$ from a predefined set $G$ as follows:
\begin{equation}
    g = \text{argmax}_{1\leq g\leq G}(\mathcal{C}(X_{\rm LLM})).
\end{equation}
In this way, the Surgical Task Dynamic Ensemble adaptively loads task-specific components: the LLM activates the corresponding LoRA parameters $\Delta W_g$ while the multi-task Q-Former selects the routine-specific memory $Q_g$.

Finally, the LLM generates the output response $y$ by processing the interleaved input sequence $X_{\rm LLM}$ constructed previously. The output response $y$ of SurgLLM is adapted using task-specific LoRA weights as follows:
\begin{equation}
    y = \text{LLM}(X_{\rm LLM}; W_0 + \Delta W_g),
\end{equation}
where $W_0$ represents the frozen weights of the base LLM, and $\Delta W_g$ denotes the task-specific LoRA weights activated by the classifier $\mathcal{C}$ for task $g$. In this way, the Surgical Task Dynamic Ensemble effectively mitigates inter-task interference while maintaining computational efficiency, enabling SurgLLM to excel across diverse surgical video understanding tasks.

\subsection{Optimization Pipeline}

We optimize SurgLLM through a two-stage progressive training strategy designed to empower robust surgical video understanding capabilities, as shown in Fig. \ref{fig1_framework}. In the first stage, we adapt the video encoder $\mathcal{E}_{v}$ to surgical scenarios through MV-Recon and video-text contrastive alignment. The masked reconstruction enables the video encoder to capture fundamental surgical visual patterns, while contrastive alignment with procedural descriptions associates visual features with surgical semantics. The optimized video encoder parameters then serve as the foundation for the visual processing of our SurgLLM. In the second stage, we optimize the multi-task Q-Former and the task-specific LoRA weights for task-specific adaptation, enabling efficient specialization across diverse surgical tasks while preventing catastrophic forgetting. As such, this progressive optimization strategy ensures SurgLLM develops from fundamental visual understanding to sophisticated multimodal reasoning, achieving effective surgical video comprehension with flexibility across diverse captioning and VQA tasks.

\begin{figure}[t]
    \centering
    \includegraphics[width=0.36\textwidth]{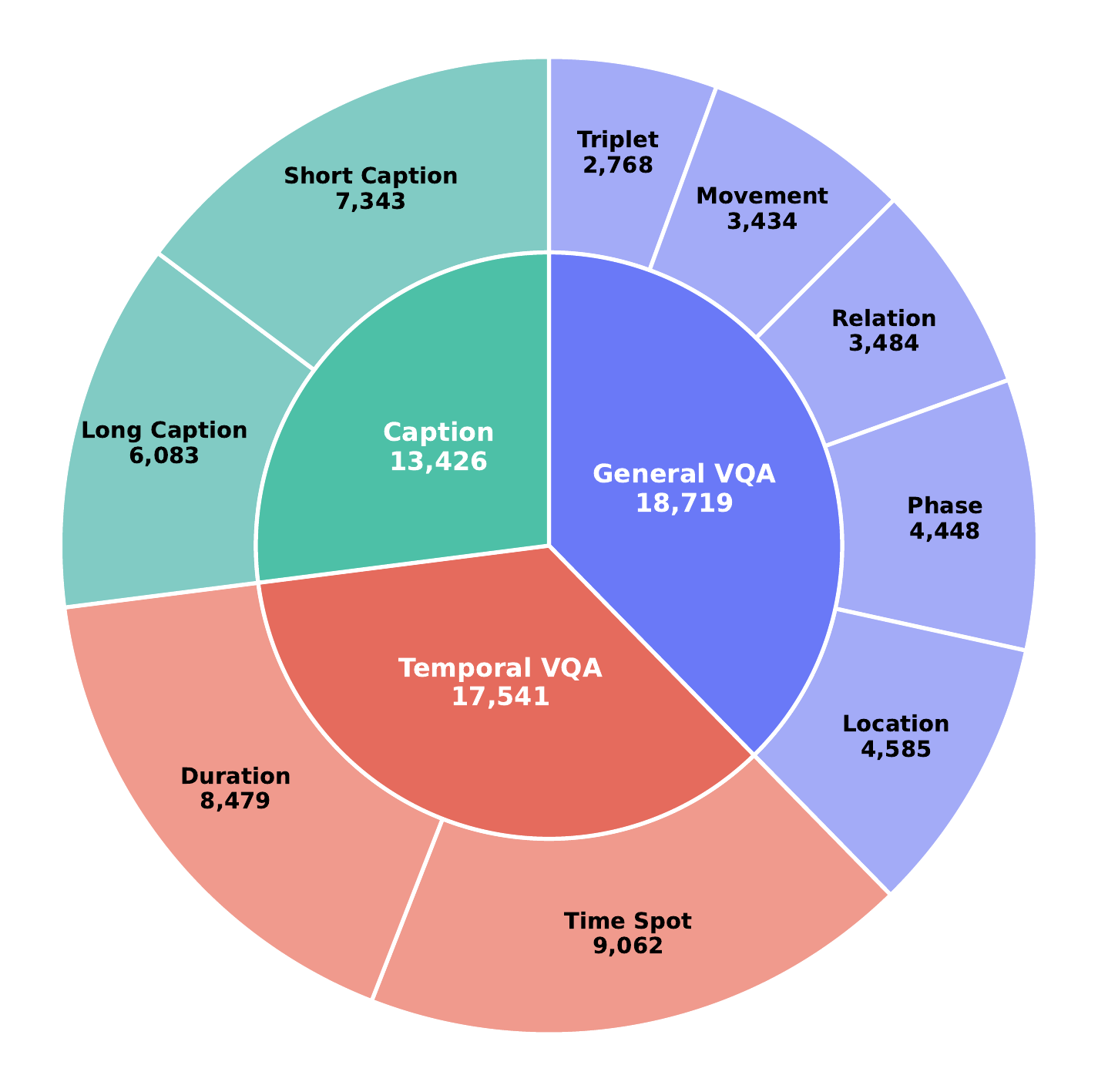}
    \caption{The distribution of the surgical video understanding dataset across three primary tasks, including the general VQA, the temporal VQA, and the caption generation. The general VQA encompasses five question types, while the temporal VQA includes both time-spot queries and event duration questions. The caption generation dataset provides both detailed long descriptions and concise short summaries of surgical video clips.}
    \label{fig6-dataset_distribution}
\end{figure}

\section{Experiments}\label{sec:experiments}

\subsection{Datasets and Implementation Details}\label{section_hyper_value}

\noindent \textbf{Surgical Video Benchmark}.
To evaluate our SurgLLM and state-of-the-art MLLMs, we build the surgical video benchmark derived from the CholecT50 dataset \cite{cholect50}. The CholecT50 dataset comprises 50 endoscopic videos of laparoscopic cholecystectomies, and provides comprehensive annotations, including surgical phases and triplets of surgical instruments, surgical actions, and operated targets. In addition, we further leverage the CholecT50-Challenge dataset \cite{cholect50challenge} with 5 surgical videos, containing bounding box annotations of surgical instruments, to benefit the preparation of instrument information. As illustrated in Fig. \ref{fig5}, we first train a surgical instrument detection model on the CholecT50-Challenge dataset, and then conduct the inference on the CholecT50 dataset to generate bounding box annotations of surgical instruments across all 50 videos. We perform the manual filtering and automatically validate these generated annotations using existing triplet annotations to ensure the accuracy of the information.

We unify the surgical video at 1 frame per second, and sort the key information of the surgical triplet, surgical phase, instrument location, and temporal information for each frame. Then, we divide the surgical videos into clips with every four frames as the basic unit for caption generation. We generate dense captions using GPT-4 \cite{gpt4} by incorporating the key information to ensure comprehensive scene description, including surgical instruments, actions, targets, absolute and relative locations, instrument movements, and surgical phases. Our dense captions include both short captions for concise scene descriptions and long captions with detailed descriptions and reasoning. For visual question-answer (VQA) pairs, we create two primary categories, including the general VQA and temporal VQA. The general VQA encompasses the tasks of {the phase recognition}, {triplet detection}, {location identification}, {relation analysis}, and {instrument movement}. The temporal VQA focuses on time-specific queries, including the procedure duration and specific time spot, validating the enhanced temporal reasoning capabilities of our SurgLLM. The distribution of our surgical video dataset is elaborated in Fig. \ref{fig6-dataset_distribution}. We randomly split the training and test sets into 80\% and 20\% at the surgical video level.

\begin{figure}[t]
    \centering
    \includegraphics[width=0.47\textwidth]{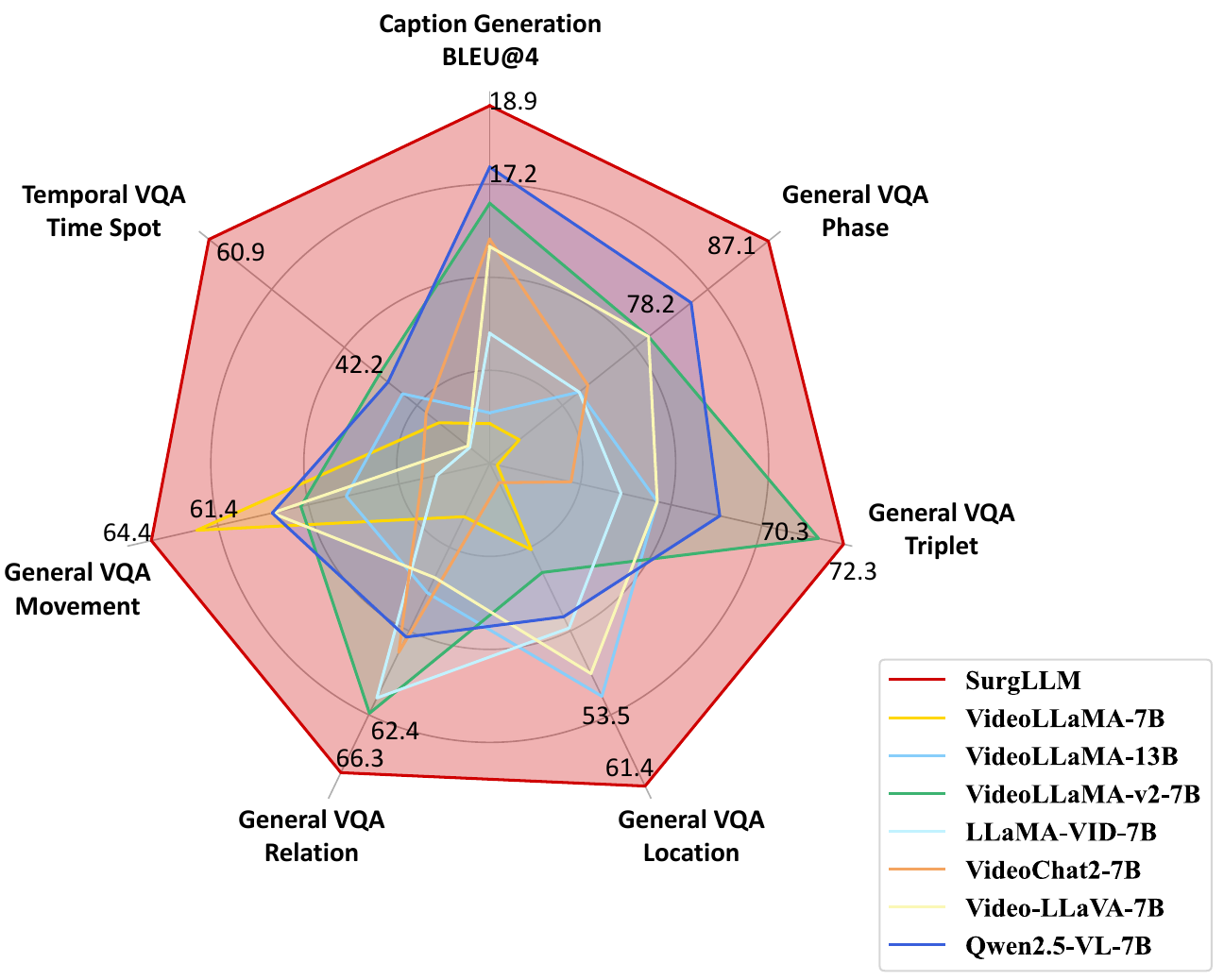}
    \caption{The radar chart comparison of our SurgLLM and the state-of-the-art video LLMs across multiple surgical video understanding dimensions. The best and second-best performances are marked. Our SurgLLM consistently outperforms existing models, particularly excelling in temporal understanding critical for surgical applications.}
    \label{fig8-radar}\vspace{-10pt}
\end{figure}

\noindent \textbf{Implementation Details}.
We implement SurgLLM and state-of-the-art MLLMs with PyTorch \cite{pytorch} on 8 NVIDIA A100 GPUs. For all models in the experiment, we unify the surgical videos into the spatial resolution of $224 \times 224$. The architecture of our SurgLLM comprises the VideoMAE \cite{videomae2} as the video encoder for Surg-Pretrain, the multi-task Q-Former \cite{instructblip} with multiple learnable memory tokens, and the Vicuna-1.5-7B \cite{vicuna} as the base LLM.

For the instrument-centric MV-Recon, we divide every 64 frames into a surgical video clip for pre-training. We randomly generate instrument-centric tube masks using the bounding boxes of surgical instruments, and set the probability $r$ as 10\% to randomly keep a small proportion of video tubes as the hint for reconstructing the masked video contents. 
We initialize the visual encoder with the weights of VideoMAE \cite{videomae}. 
We adopt Adam to optimize the video encoder until convergence with the learning rate of $5\times10^{-4}$. We implement multi-scale temporal masking with varying durations of 2, 4, 8, and 16 frames to capture fine-grained temporal patterns. For surgical video context alignment, we adopt AdamW optimizer with the learning rate of $1\times 10^{-5}$ and the weight decay of $0.02$. We perform the multimodal contrastive training with the short caption training data for 3 epochs. For TM-Tuning, we set up the structure of SurgLLM. We first fine-tune the multi-task Q-Former with full parameters while freezing LLM parameters on long caption training data for 3 epochs, using the learning rate of $5\times 10^{-4}$ and the weight decay of $0.05$. After that, we further fine-tune both the multi-task Q-Former and LLM using task-specific LoRAs with the rank of 8, the alpha of 8 in the learning rate of $1\times 10^{-4}$ for 3 epochs on VQA training data, where we adopt the surgical task dynamic ensemble in the fine-tuning.

\begin{table*}[t]
\centering
\caption{Comparison of SurgLLM and state-of-the-art video LLMs on surgical video caption generation.}
\scriptsize
\setlength{\tabcolsep}{15pt}
\centering
\begin{tabular}{l|ccccccc}
\toprule
Methods & BLEU@1 & BLEU@2 & BLEU@3 & BLEU@4 & CIDEr & ROUGE-L & METEOR \\
\midrule
VideoLLaMA-7B \cite{videollama} &  45.2 &27.4&16.3&10.1& 5.7     & 19.0    & 26.2   \\ 

VideoLLaMA-13B  \cite{videollama} &  45.9  &27.9&16.7&10.4  & 4.8      & 19.0       & 26.8   \\ 
LLaMA-VID-7B \cite{llamavid}  & 45.1&28.6&18.4&12.6 & 9.8     & 19.5   & 28.7    \\

Video-LLaVA-7B  \cite{videollava} &  49.4 &31.8&21.2&15.0 & 10.0   & 20.7 & 31.7   \\

VideoLLaMA-v2-7B \cite{videollamav2} & 52.5 &34.2&22.8& 16.2 & 15.0    & 22.4   & 21.9    \\

VideoChat2-7B \cite{videochat}&50.0 &33.4&22.1&15.2 & 11.7   & 22.6    & 22.8  \\ 

Qwen2.5-VL-7B \cite{Qwen2.5-VL} &45.3 & 31.9& 22.7&17.2&12.0&21.5&21.4\\

\rowcolor[rgb]{ .949,  .949,  .949} 
\textbf{SurgLLM (Ours)}  &  \textbf{55.0}&\textbf{37.6}&\textbf{26.0}&\textbf{18.9}  & \textbf{17.5}   & \textbf{23.0}   & \textbf{36.0}    \\ \bottomrule
\end{tabular}
\label{comparison-caption-generation}
\end{table*}

\begin{figure*}[t]
    \centering
    \includegraphics[width=1\textwidth]{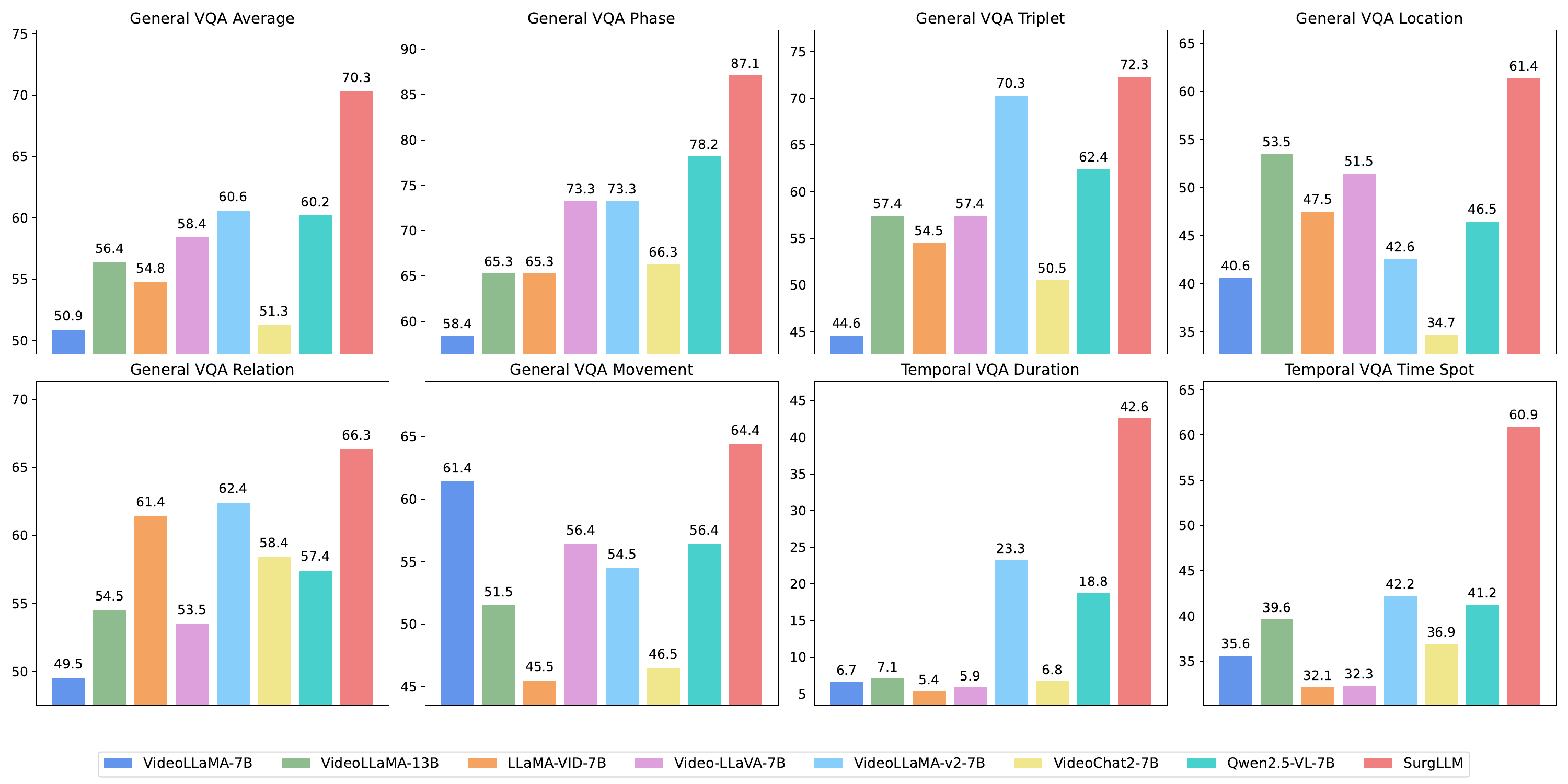}
    \caption{Comparison of SurgLLM and state-of-the-art video LLMs across surgical VQA tasks. The general VQA tasks include phase recognition, triplet detection, location identification, relation analysis, movement analysis, and their average score. The temporal VQA tasks include the IoU score and the accuracy for duration and time spot queries, respectively.}
    \label{fig7-comparison-VQA}
\end{figure*}

\noindent \textbf{Evaluation Metrics}.
We employ comprehensive evaluation metrics tailored to different surgical video task requirements. For the caption generation task, we utilize diverse natural language metrics, including the BLEU \cite{papineni2002bleu}, CIDEr \cite{vedantam2015cider}, ROUGE-L \cite{banerjee2005meteor}, and METEOR \cite{banerjee2005meteor} scores. For the general VQA tasks, we employ GPT-4 \cite{gpt4} to validate the correctness of the prediction given the ground truth, and then calculate the accuracy regarding diverse types of queries, including phase recognition, triplet detection, location identification, relation analysis, and movement analysis. For the temporal VQA tasks, we calculate the Intersection over Union (IoU) score for the duration prediction to measure the overlap between predicted and ground-truth time periods, and compute the accuracy of time spot prediction with the ground truth. In this way, these metrics can reflect the general and temporal understanding capabilities of SurgLLM and other video LLMs on surgical videos.

\subsection{Comparison with State-of-the-Art Video LLMs}
We conduct comprehensive comparisons across three fundamental tasks of surgical videos, including caption generation, general VQA, and temporal VQA. As illustrated in Fig. \ref{fig8-radar}, these evaluations demonstrate the superior performance of SurgLLM in addressing the challenges of surgical video understanding that general-purpose video LLMs fail to handle effectively.

\begin{table*}[t]
\centering
\caption{Comprehensive ablation study of different components in the SurgLLM framework.}
\scriptsize

\begin{minipage}{1.0\textwidth}
\centering
\subcaption{Ablation study on the surgical video caption generation task.}
\setlength{\tabcolsep}{12pt}
\begin{tabular}{cc|cccccccc}
\toprule
MV-Recon & Multi-task Q-Former & BLEU@1 & BLEU@2 & BLEU@3 & BLEU@4 & CIDEr & ROUGE-L & METEOR \\
\midrule
 &  & 43.5 & 29.0 & 20.1 & 13.7 & 14.4 & 19.6 & 29.0 \\
 
 $\checkmark$ &  & 49.3 & 31.8 & 21.1 & 14.9 & 15.9 & 20.6 & 30.5 \\
 & $\checkmark$ & 50.8 & 31.8 & 22.7 & 16.2 & 16.9 & 21.7 & 32.4 \\
 \rowcolor[rgb]{ .949,  .949,  .949} 
 $\checkmark$ & $\checkmark$ & \textbf{55.0} & \textbf{37.6} & \textbf{26.0} & \textbf{18.9} & \textbf{17.5} & \textbf{23.0} & \textbf{36.0} \\
\bottomrule
\end{tabular}
\vspace{5pt}
\end{minipage}


\begin{minipage}{1.0\textwidth}
\centering
\subcaption{Ablation study on the general VQA and temporal VQA tasks with different multi-task Q-Former configurations.}
\setlength{\tabcolsep}{8pt}
\begin{tabular}{ccc|cccccc|cc}
\toprule
\multirow{2}{*}{MV-Recon} & \multirow{2}{*}{\makecell{Task Dynamic \\ Ensemble}}& \multirow{2}{*}{\makecell{Multi-task \\ Q-Former}} & \multicolumn{6}{c}{General VQA} & \multicolumn{2}{|c}{Temporal VQA}  \\ \cmidrule(lr){4-9} \cmidrule(lr){10-11}
&& &Phase & Triplet & Location & Relation & Move. & Avg &Duration&Time Spot \\
\midrule
& & & 73.3 & 53.5 & 50.5 & 49.5 & 57.4 & 56.8 & 33.1 & 44.8 \\
$\checkmark$ & &  & 67.3 & 63.4 & 52.4 & 58.4 & 58.4 & 60.0 & 35.5 & 48.7 \\
 & $\checkmark$ & &  75.3 & 64.4 & 51.5 & 62.4 & 56.4 & 62.0 & 34.2 & 45.5 \\
$\checkmark$ & $\checkmark$ & & 76.2 & 66.3 & 52.5 & 59.4 & 61.4 & 63.2 & 36.6 & 52.2 \\
 
 &  & I-QFormer & 69.3 & 56.4 & 51.5 & 54.5 & 60.4 & 58.4 & 34.9 & 48.4 \\
$\checkmark$ &  & I-QFormer & 68.3 & 65.3 & 51.5 & 57.4 & 58.4 & 60.2 & 35.7 & 49.2 \\
 & $\checkmark$ & I-QFormer & 77.2 & 71.3 & 56.4 & 63.4 & 62.4 & 66.1 & 35.3 & 47.5 \\
$\checkmark$ & $\checkmark$ & S-QFormer & 78.2 & 59.4 & 50.5 & 57.4 & 62.4 & 61.6 & 38.2 & 54.4 \\
\rowcolor[rgb]{ .949,  .949,  .949} 
$\checkmark$ & $\checkmark$ & I-QFormer & \textbf{87.1} & \textbf{72.3} & \textbf{61.4} & \textbf{66.3} & \textbf{64.4} & \textbf{70.3} & \textbf{42.6} & \textbf{60.9} \\
\bottomrule
\end{tabular}
\end{minipage}
\label{ablation}
\end{table*}

\subsubsection{Comparison on Caption Generation}
The caption generation task requires the model to generate an illustration of surgical videos. As shown in Table \ref{comparison-caption-generation}, our SurgLLM demonstrates superior performance across all metrics of captioning, particularly the BLEU@4 of 18.9\% and the METEOR of 36.0\%. The significant improvements demonstrate the enhanced understanding of our SurgLLM on surgical scene dynamics and semantic relationships, which benefits from improved alignment between visual surgical content and textual representations.

\subsubsection{Comparison on General VQA}
We further validate the effectiveness of our SurgLLM and state-of-the-art video LLMs on comprehensive surgical VQA tasks. As elaborated in Fig. \ref{fig7-comparison-VQA}, SurgLLM achieves superior accuracy across all general VQA tasks, with an average improvement of 9.7\% over the second-best method \cite{videollamav2}. The comparative analyses provide compelling evidence that our SurgLLM not only excels in instruction following but also demonstrates enhanced comprehension of video content across multiple dimensions of surgical tasks. These findings collectively affirm the efficacy of our SurgLLM in the context of surgical video understanding and question-answering, positioning it as a state-of-the-art solution in this domain.


\subsubsection{Comparison on Temporal VQA}
Furthermore, we evaluate the temporal understanding capabilities of our SurgLLM and the state-of-the-art video LLMs through temporal VQA tasks. These temporal VQA samples are tailored to probe the perception and reasoning about temporal relationships within surgical videos.
Fig. \ref{fig7-comparison-VQA} reveals that existing video LLMs exhibit limited temporal perception capabilities, lacking specialized modeling for surgical video temporal characteristics. In contrast, our SurgLLM, enhanced with TM-Tuning, achieves the best performance of 60.9\% in time spot and 42.6\% in duration, and obtains the substantial improvements of 18.7\% in time spot and 19.3\% in duration compared to the second-best Video-LLaMA-v2-7B \cite{videollamav2}, respectively. These comparisons demonstrate the effectiveness of our SurgLLM in precise temporal reasoning for surgical video understanding.

\subsection{Ablation Study}

To validate the effectiveness of each component in our SurgLLM, we conduct systematic ablation studies, focusing on key innovations, including MV-Recon component, multi-task Q-Former, and surgical task dynamic ensemble.

\noindent \textbf{Ablation Study on MV-Recon}.
Table \ref{ablation} (a) demonstrates the consistent improvements of our instrument-centric MV-Recon across all evaluation dimensions. The MV-Recon component addresses the challenge of inadequate visual content perception in surgical videos by capturing unique foreground-background dynamics. The caption generation results show improvements of 4.2\%, 5.8\%, 3.3\%, and 2.7\% in BLEU metrics with varying n-gram, with gains of 0.6\%, 1.3\%, and 3.6\% in CIDEr, ROUGE-L, and METEOR, respectively. For VQA tasks, we observe remarkable performance gains of 3.2\% in general VQA and 3.9\% in temporal VQA, along with 2.4\% improvement in duration, validating the effectiveness of our surgical-specific masking strategies.

\noindent \textbf{Ablation Study on Surgical Task Dynamic Ensemble}.
The Surgical Task Dynamic Ensemble addresses the challenge of task diversity while preventing catastrophic forgetting. Results in Table \ref{ablation} (b) demonstrate the average accuracy improvements ranging from 56.8\% to 62.0\%, with particularly notable enhancements in the triplet task with 10.9\% and the relation task with 12.9\%, validating the effectiveness of our ensemble approach in handling multifaceted surgical procedures.

\noindent \textbf{Ablation Study on Multi-task Q-Former Designs}.
Table \ref{ablation} (b) demonstrates the significant impact of multi-task Q-Former on surgical video understanding, which transfers valuable visual perception capabilities to the surgical video domain. Moreover, we investigate the implementations of multi-task Q-Former, where the independent Q-Former refers to the Q-Former being randomly initialized for each task (denoted as I-QFormer), and the shared Q-Former refers to the Q-Former weights being shared and trained across all tasks (denoted as S-QFormer). For general VQA, the independent Q-Former improves average performance by 8.7\% compared to the shared Q-Former. For temporal VQA, the independent Q-Former consistently outperforms the shared Q-Former with 4.4\% in duration and 6.5\% in time spot.

\noindent \textbf{Ablation Study on Temporal-aware Embedding Strategies}.
Table \ref{ablation study of comparing} investigates different temporal embedding approaches, addressing the temporal awareness capabilities of our SurgLLM framework. The direct-embedding approach consistently underperformed due to abrupt visual-textual information juxtaposition without temporal integration. Front-embedding demonstrated limitations with long visual token sequences. Our textual-visual temporal interleave embeddings strategy emerged as most effective, achieving 14.8\% improvement in time spot of temporal VQA compared to the second-best method. This validates the critical role of our temporal interleaving approach in enabling precise temporal reasoning for surgical video understanding.

\begin{table*}[t]
\centering
\caption{Ablation study of temporal embedding designs in TM-Tuning.}
\scriptsize
\setlength{\tabcolsep}{12pt}
\begin{tabular}{l|cccccc|cc}
\toprule
\multirow{2}{*}{\textbf{Embedding Strategy}}& \multicolumn{6}{c}{General VQA}& \multicolumn{2}{|c}{Temporal VQA} 
\\ \cmidrule(lr){2-7} \cmidrule(lr){8-9}
&Phase & Triplet & Location & Relation & Move. & Avg &Duration&Time Spot  \\ 
\midrule
 Direct Embedding & 76.2& 60.4 &60.4 &55.5 &61.6&62.8& 31.2& 40.2 \\
Front Embedding & 81.2& 70.3 & 50.5&57.4& 59.4 & 63.8& 34.8&46.1 \\
\rowcolor[rgb]{ .949,  .949,  .949} 
Interleave Embedding (Ours) & \textbf{87.1}& \textbf{72.3} & \textbf{61.4} &\textbf{66.3}&\textbf{64.4}&\textbf{70.3}& \textbf{42.6} & \textbf{60.9}  \\
\bottomrule
\end{tabular}
\label{ablation study of comparing}
\end{table*}

\begin{table*}[t]
\centering
\caption{Impact of video segment length on SurgLLM performance across different surgical video tasks.}\label{length-ablation}
\scriptsize
\begin{minipage}{1.0\textwidth}
\subcaption{Impact of video segment length on caption generation.}
\centering
\setlength{\tabcolsep}{16pt}
\begin{tabular}{c|ccccccc}
\toprule
\multicolumn{1}{c|}{Length} & \multicolumn{1}{c}{BLEU@1} & {BLEU@2} &{BLEU@3} &{BLEU@4} &\multicolumn{1}{c}{CIDEr} & \multicolumn{1}{c}{ROUGE-L} & \multicolumn{1}{c}{METEOR} \\ \midrule
2 &  \textbf{55.6} &37.3&25.8&18.2& 17.0  & 21.9   & 32.5  \\ 
4  &  55.0 &\textbf{37.6}&\textbf{26.0}&\textbf{18.9}& \textbf{17.5} & \textbf{23.0}     & \textbf{36.0} \\ 
8 & 54.2&36.9&25.1& 17.9& 16.9   & 22.4   & 34.0    \\ 
16  & 54.8&37.1&25.5&17.4 & 17.2     & 21.8  &32.5 \\ \bottomrule
\end{tabular}
\vspace{5pt}
\end{minipage}
\begin{minipage}{1.0\textwidth}
\centering
\subcaption{Impact of video segment length on general VQA and temporal VQA.}
\setlength{\tabcolsep}{15pt}
\begin{tabular}{c|cccccc|cc}
\toprule
\multirow{2}{*}{{Length}} & \multicolumn{6}{c}{General VQA}& \multicolumn{2}{|c}{Temporal VQA}\\ \cmidrule(lr){2-7} \cmidrule(lr){8-9}
&Phase & Triplet & Location & Relation & Move. & Avg &Duration&Time Spot \\ \midrule
2 & 85.3 & 70.4 & 59.5 & 65.1 & 62.5 & 68.6 & 40.7 & 58.8\\
4 & \textbf{87.1} & \textbf{72.3} & \textbf{61.4} & \textbf{66.3} & \textbf{64.4} & \textbf{70.3}& \textbf{42.6} & \textbf{60.9}\\
8 & 86.2 & 71.4 & 60.4 & 65.5 & 63.6 & 69.4 & 41.3 & 59.2\\
16 & 84.2 & 70.3 & 59.3 & 64.6 & 62.1 & 68.1 & 40.1 & 58.1\\ \bottomrule
\end{tabular}
\end{minipage}
\end{table*}

\subsection{Hyperparameter Analysis}

\textbf{Analysis of Video Segment Duration}.
To investigate the robustness of our SurgLLM framework across different temporal granularities, we evaluate performance with varying video segment durations, as shown in Table \ref{length-ablation}. Overall, our SurgLLM demonstrates relatively robust performance across different segment lengths, with moderate durations yielding optimal results. The experimental results reveal that 4-second segments consistently achieve the best performance across both caption generation and VQA tasks, with BLEU@4 of 18.9\% for captioning and 70.3\% average accuracy for general VQA. Shorter segments (\textit{e.g.}, 2 seconds) show competitive but slightly inferior performance, likely due to insufficient temporal context for capturing complete surgical actions and their sequential dependencies. Conversely, longer segments (\textit{e.g.}, 16 seconds) exhibit modest performance degradation, suggesting that extended temporal windows may introduce irrelevant information that challenges the model's ability to focus on critical surgical events. Notably, the performance variations across different segment lengths remain relatively modest. For instance, in caption generation, BLEU@4 scores range from 17.4\% to 18.9\%, while the average accuracy of general VQA varies between 68.1\% and 70.3\%. This limited variation demonstrates the robustness of our temporal-aware design and TM-Tuning in handling diverse temporal granularities. These findings confirm that while temporal granularity does influence performance, our SurgLLM framework maintains stable and effective surgical video understanding capabilities across a reasonable range of segment lengths, highlighting the practical applicability of our SurgLLM framework in real-world surgical scenarios where temporal segmentation may vary.


\begin{figure}[t]
    \centering
    \includegraphics[width=0.42\textwidth]{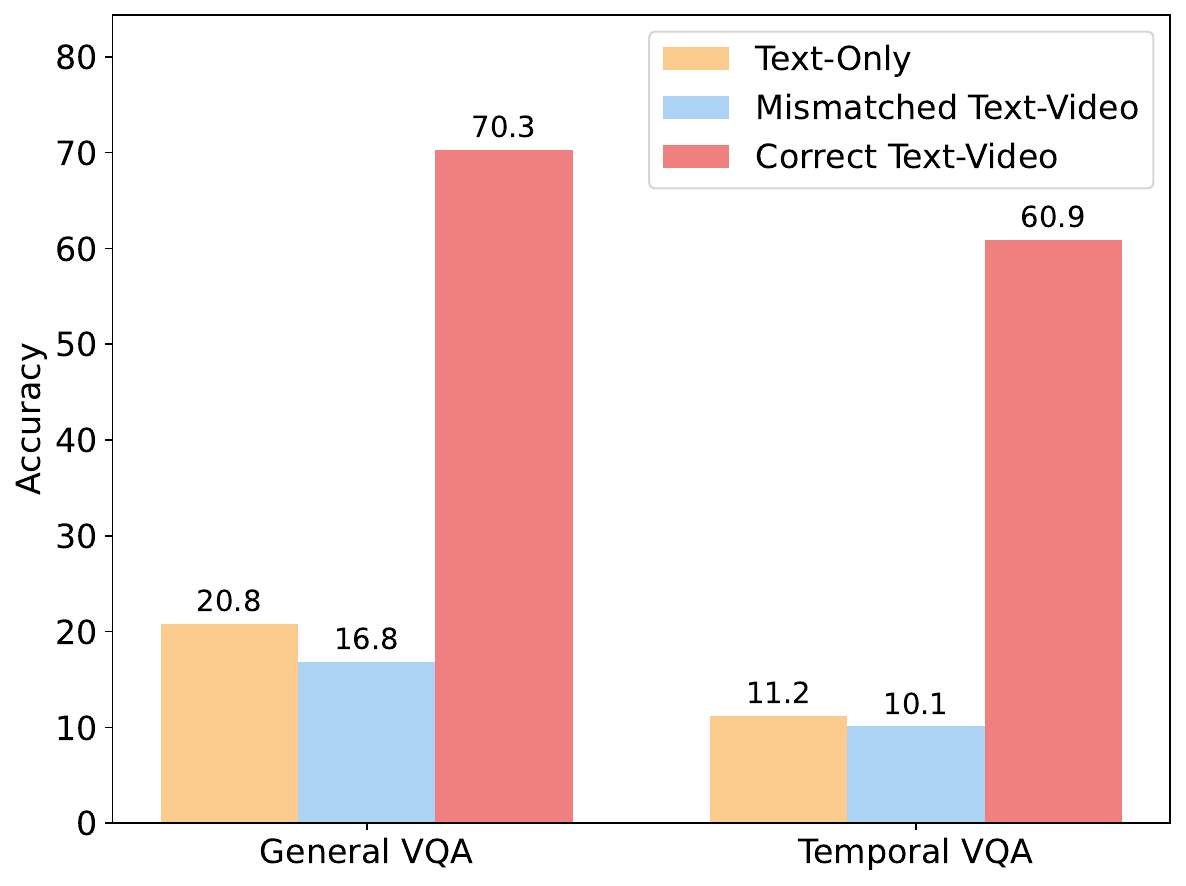}
    \caption{Analysis of input demonstrates that surgical video understanding benefits from correct text and video inputs.}
    \label{fig9-visual_input}
\end{figure}

\noindent \textbf{Analysis of Visual Input Necessity}.
We further validate the impact of visual-dependent input quality on the performance. We implement the inference of SurgLLM in three scenarios, including the text-only input, the mismatched text-video pair, and the correct text-video pair. As illustrated in Fig. \ref{fig9-visual_input}, compared with the baseline of text-only input, the mismatched text-video pair degrades the performance by providing misleading information. Given the correct text-video pair, SurgLLM benefits from the capability to perceive and respond based on visual surgical content and demonstrates significant improvements compared to the text-only and mismatched video-text baselines, by a 49.5\% and 53.5\% improvement in the average score of general VQA, and a 49.7\% and 50.8\% improvement in the time spot of temporal VQA, respectively. In this way, these results confirm that our SurgLLM effectively utilizes visual context for accurate responses, validating the necessity of multimodal understanding in surgical video understanding.


\section{Conclusion}\label{sec:conclusion}
In this paper, we present SurgLLM, a versatile multimodal large language model framework specifically designed for comprehensive surgical video understanding. Our SurgLLM framework addresses the critical limitations of existing video LLMs in surgical scenarios through three key innovations, including Surg-Pretrain with multi-scale instrument-centric masked video reconstruction to capture surgical dynamics, TM-Tuning with textual-visual temporal interleave embeddings for precise temporal reasoning, and Surgical Task Dynamic Ensemble for efficient multi-task adaptation. Extensive experiments demonstrate significant improvements over state-of-the-art methods across caption generation, general VQA, and temporal VQA tasks, with particularly notable gains in temporal duration and time spot tasks, validating the effectiveness of SurgLLM as a unified solution for computer-assisted surgery and establishing a concrete foundation for comprehensive surgical video analysis.



\section*{Acknowledgments}
This work was supported in part by the InnoHK Program of the Hong Kong SAR Government and the National Natural Science Foundation of China (Grant No. \#62306313).


\bibliographystyle{IEEEtran}
\bibliography{czmybibliography}

\newpage

\begin{figure*}[!t]
    \centering
    \includegraphics[width=0.95\textwidth]{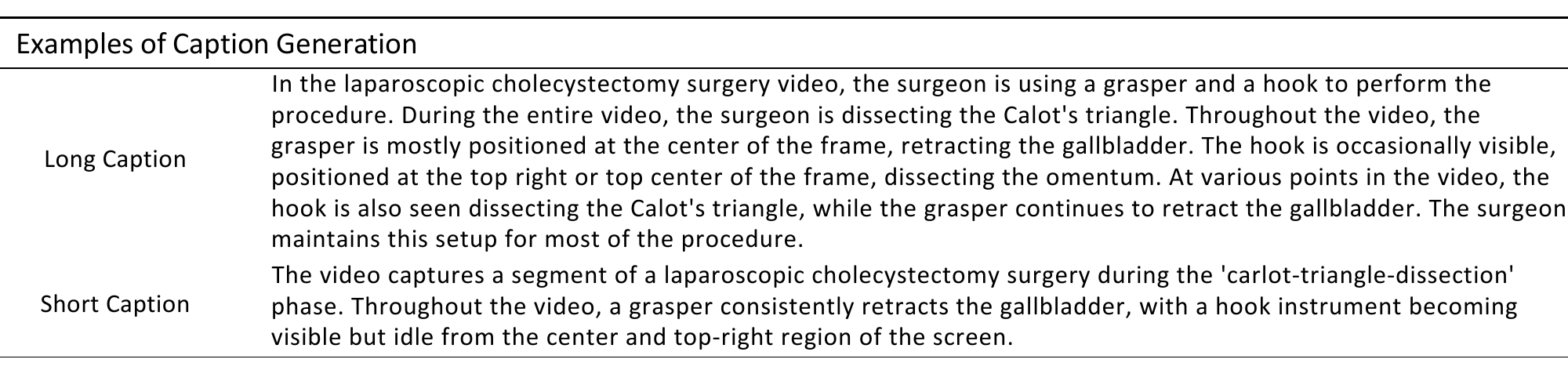}
    \caption{{Examples of caption generation in our processed CholecT50 dataset.}}
    \label{fig_s1}
\end{figure*}

\begin{figure*}[!t]
    \centering
        \includegraphics[width=0.95\textwidth]{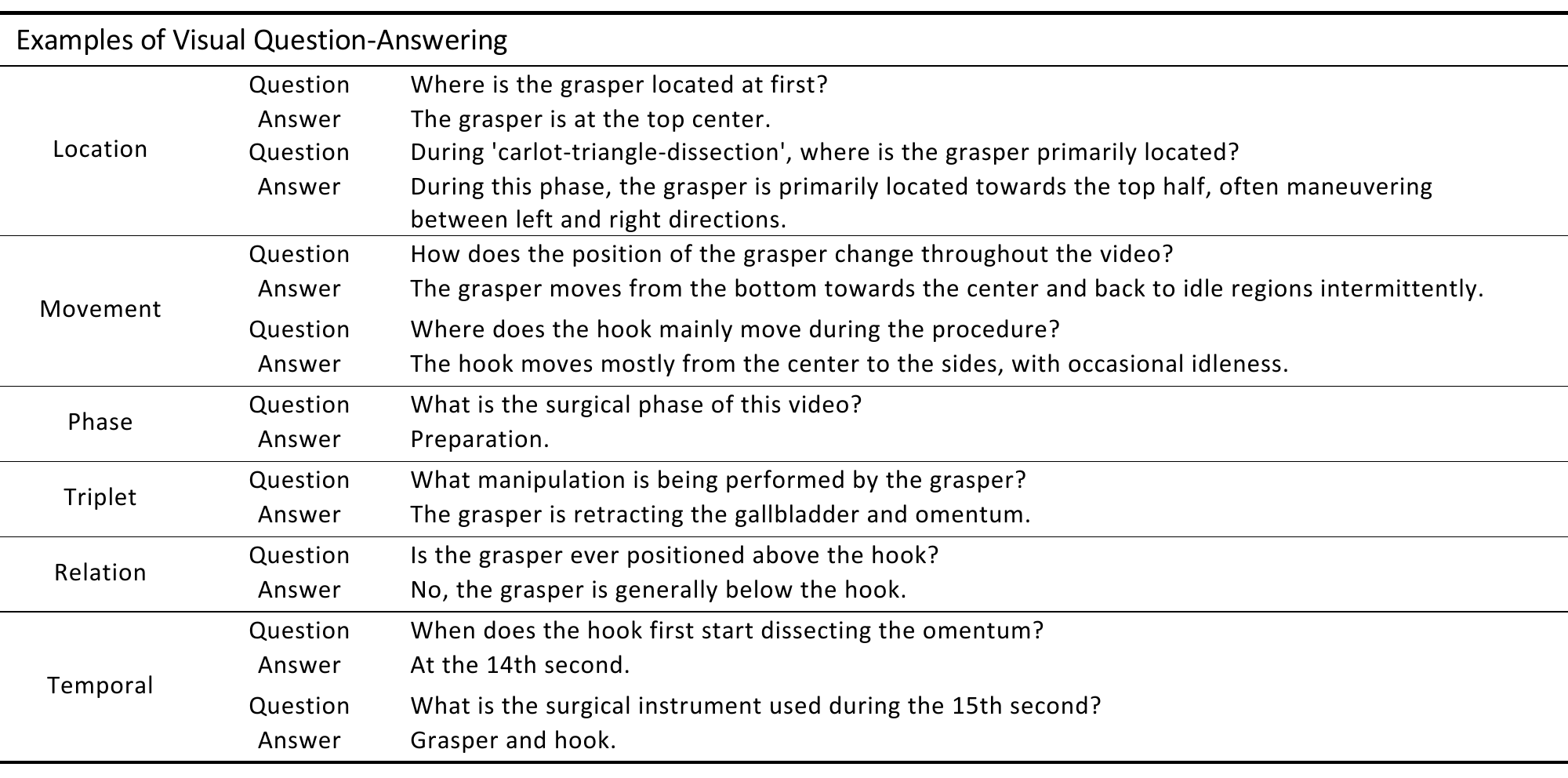}
    \caption{{Examples of general VQA and temporal VQA in our processed CholecT50 dataset.}}
    \label{fig_s2}
\end{figure*}

\begin{figure*}[!t]
    \centering
    \includegraphics[width=0.90\textwidth]{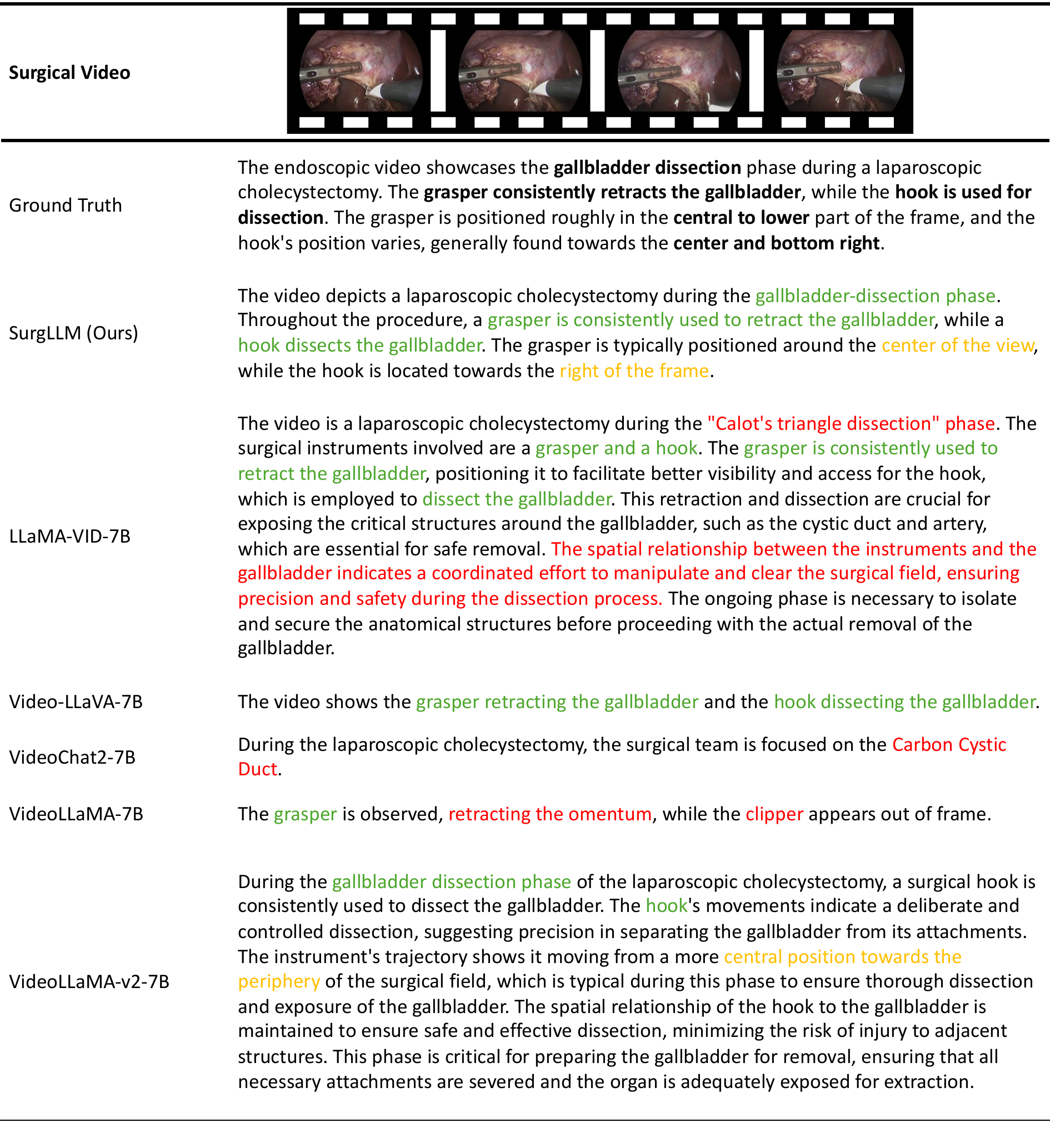}
    \caption{{Comparison of SurgLLM and state-of-the-art Video LLMs on the caption generation task. A green part refers to a perfect match, a yellow part refers to a partial match, and a red part refers to a mismatch.}}
    \label{fig_s3}
\end{figure*}

\begin{figure*}[h]
    \centering
    \includegraphics[width=0.9\textwidth]{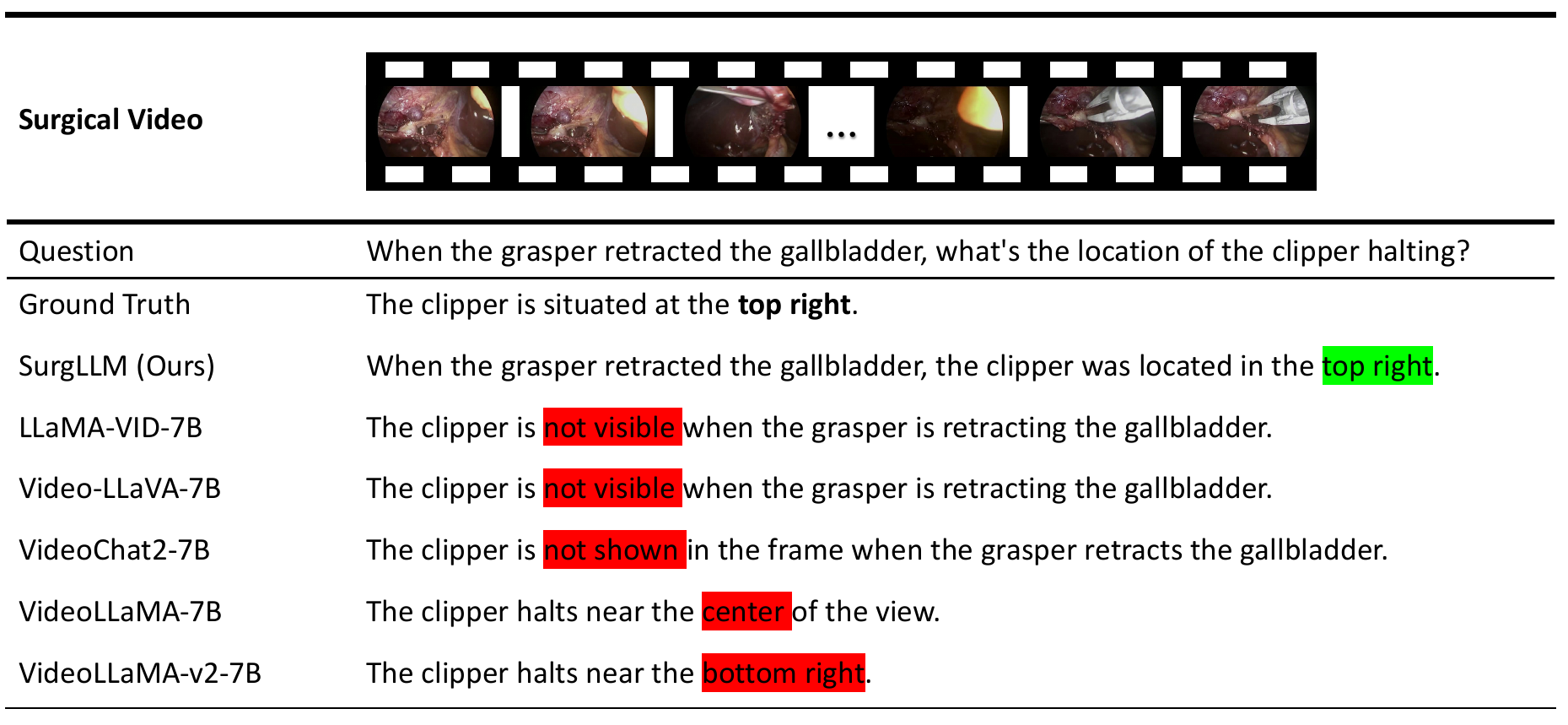}
    \caption{{Comparison of SurgLLM and state-of-the-art Video LLMs on the general VQA regarding the location task. A green part refers to a perfect match, and a red part refers to a mismatch.}}
    \label{fig_s4}
\end{figure*}

\begin{figure*}[h]
    \centering
    \includegraphics[width=0.9\textwidth]{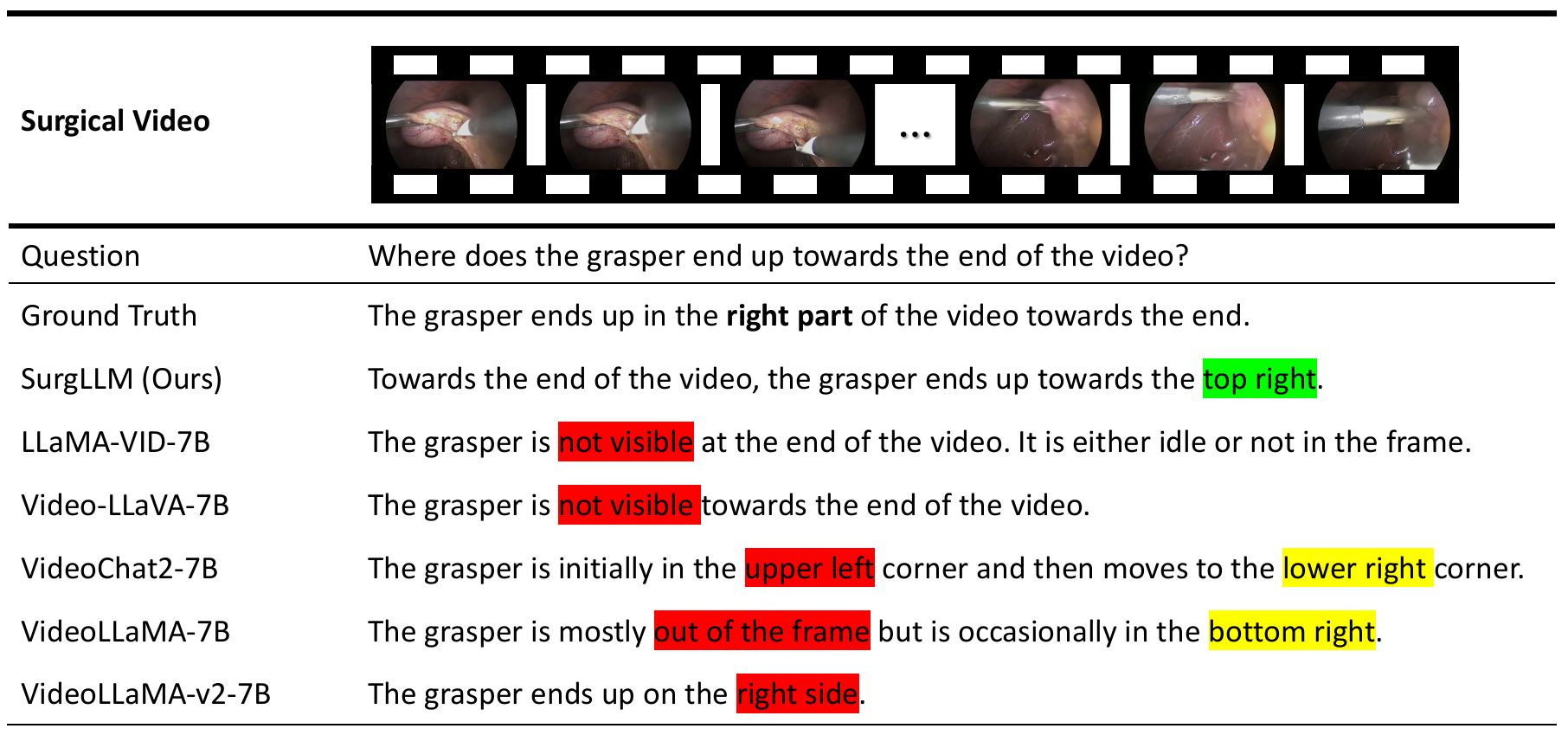}
    \caption{{Comparison of SurgLLM and state-of-the-art Video LLMs on the general VQA regarding the movement task. A green part refers to a perfect match, a yellow part refers to a partial match, and a red part refers to a mismatch.}}
    \label{fig_s5}
\end{figure*}

\begin{figure*}[h]
    \centering
    \includegraphics[width=0.9\textwidth]{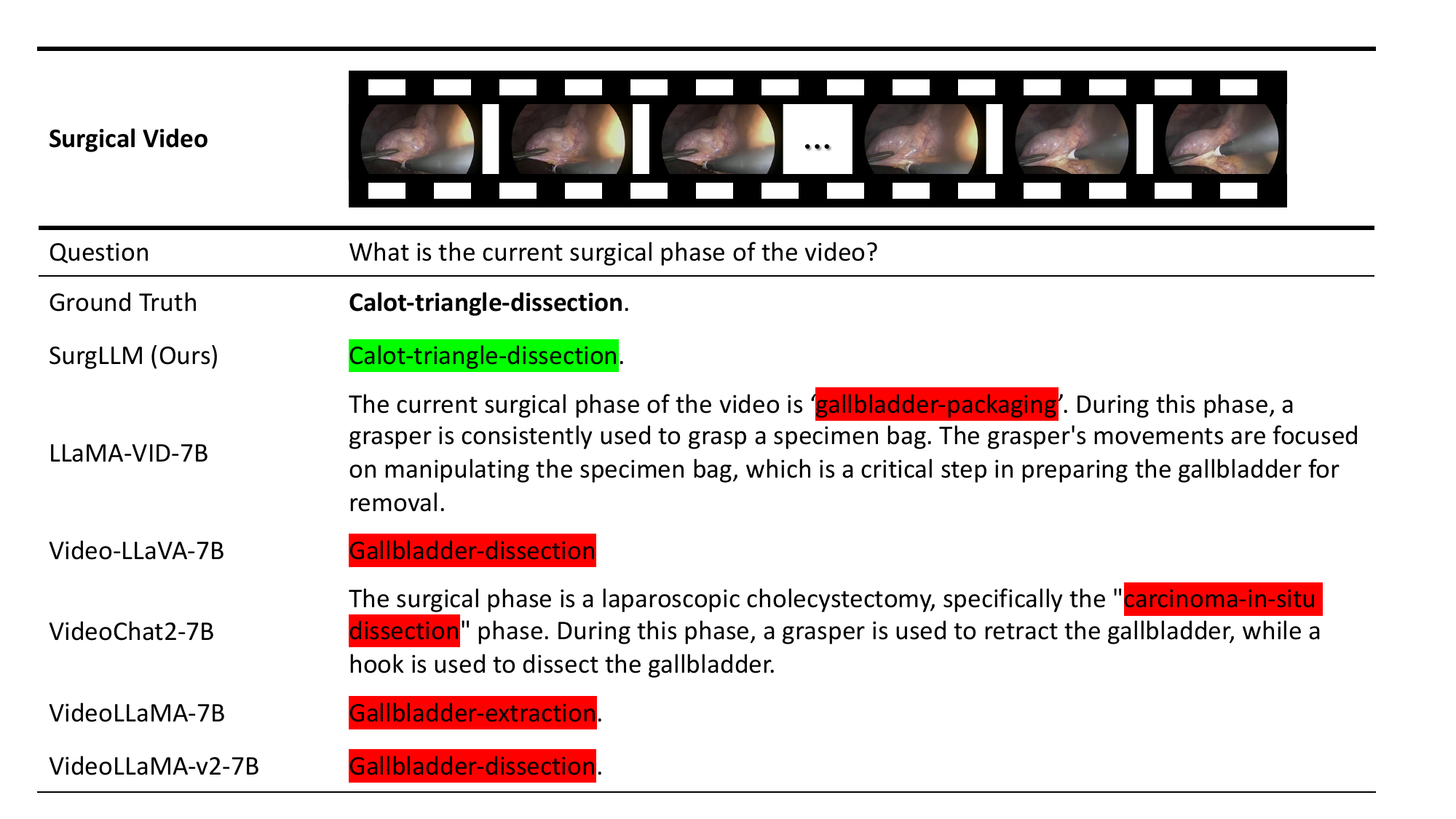}
    \caption{{Comparison of SurgLLM and state-of-the-art Video LLMs on the general VQA regarding the phase task. A green part refers to a perfect match, and a red part refers to a mismatch.}}
    \label{fig_s6}
\end{figure*}

\begin{figure*}[h]
    \centering
    \includegraphics[width=0.9\textwidth]{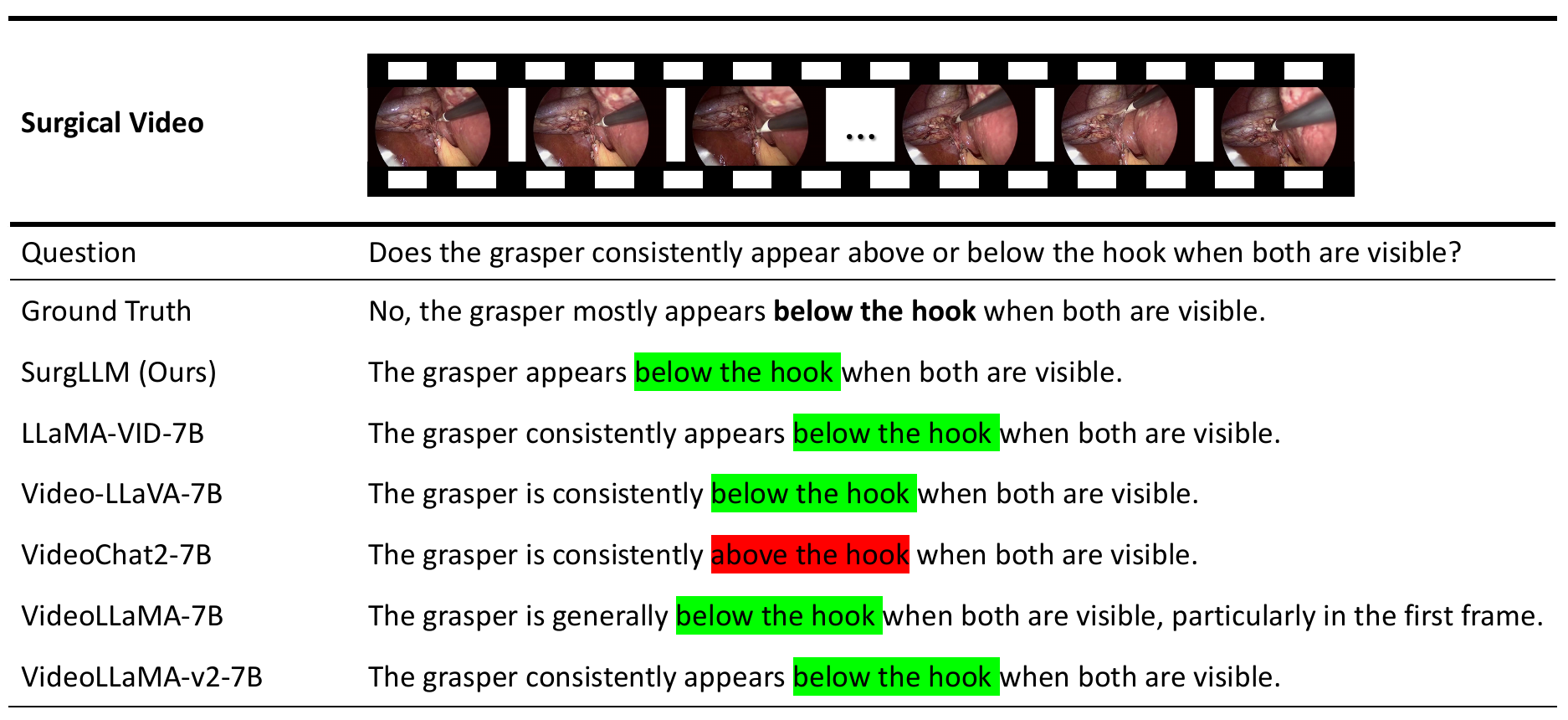}
    \caption{{Comparison of SurgLLM and state-of-the-art Video LLMs on the general VQA regarding the relation task. A green part refers to a perfect match, and a red part refers to a mismatch.}}
    \label{fig_s7}
\end{figure*}

\begin{figure*}[h]
    \centering
    \includegraphics[width=0.9\textwidth]{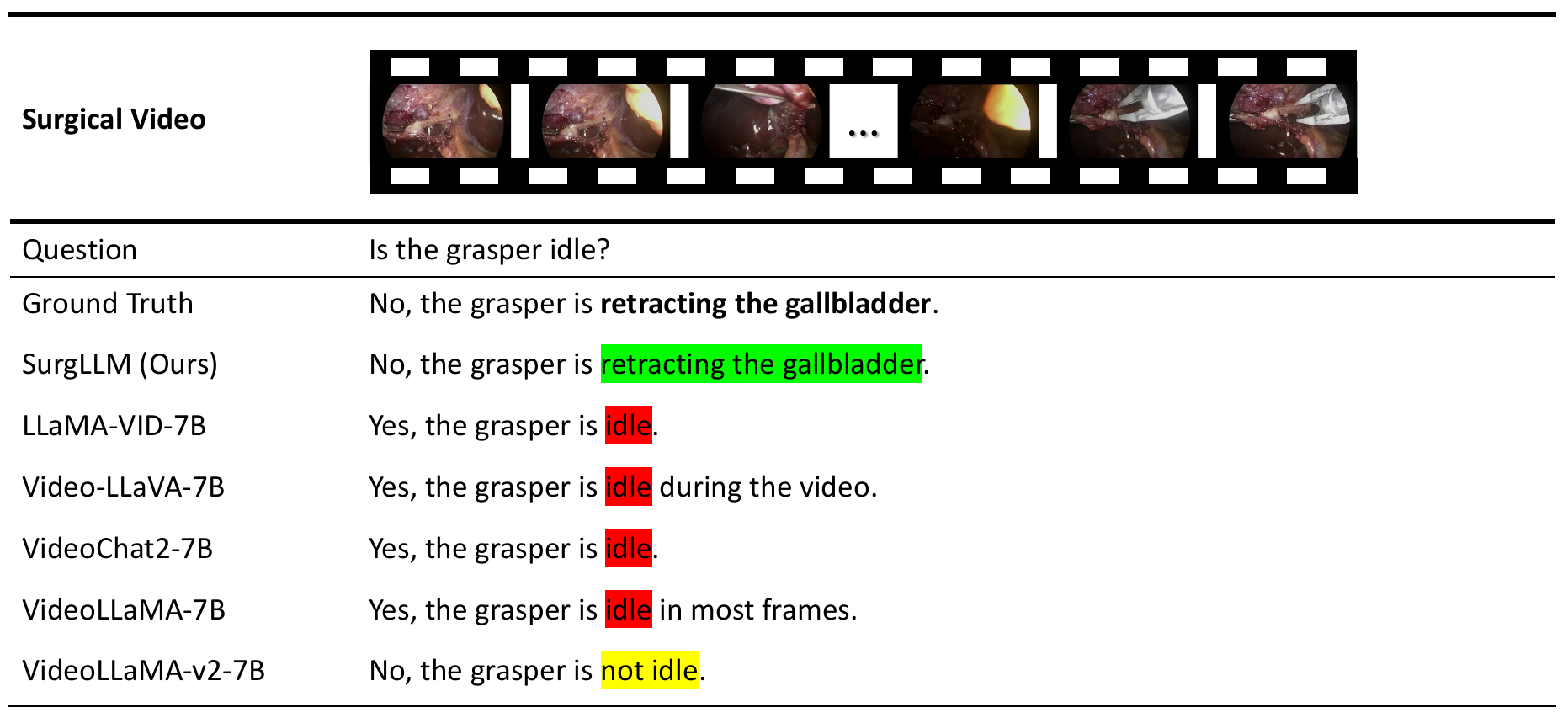}
    \caption{{Comparison of SurgLLM and state-of-the-art Video LLMs on the general VQA regarding the triplet task. A green part refers to a perfect match, a yellow part refers to a partial match, and a red part refers to a mismatch.}}
    \label{fig_s8}
\end{figure*}

\begin{figure*}[h]
    \centering
    \includegraphics[width=0.9\textwidth]{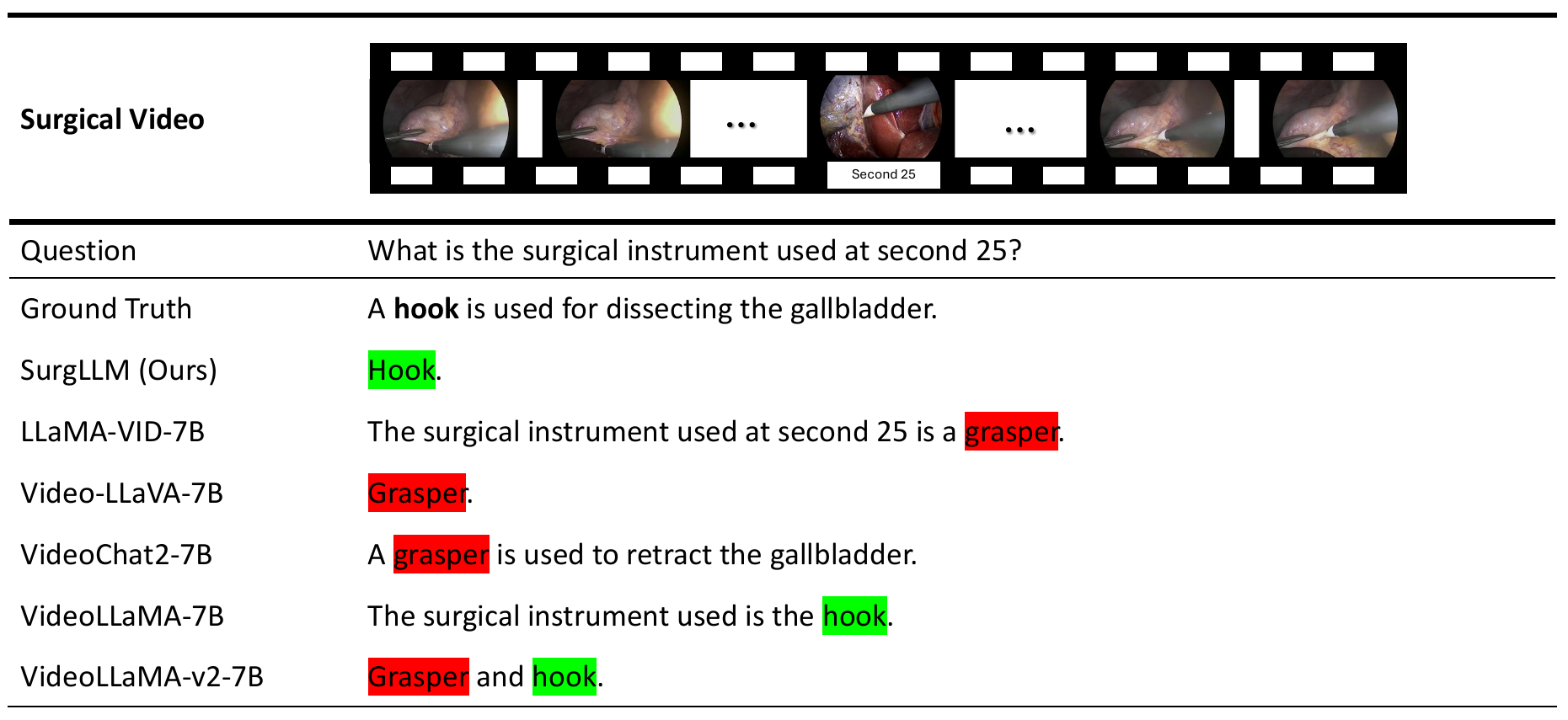}
    \caption{{Comparison of SurgLLM and state-of-the-art Video LLMs on the temporal VQA regarding the time spot task. A green part refers to a perfect match, and a red part refers to a mismatch.}}
    \label{fig_s9}
\end{figure*}

\begin{figure*}[h]
    \centering
    \includegraphics[width=1\textwidth]{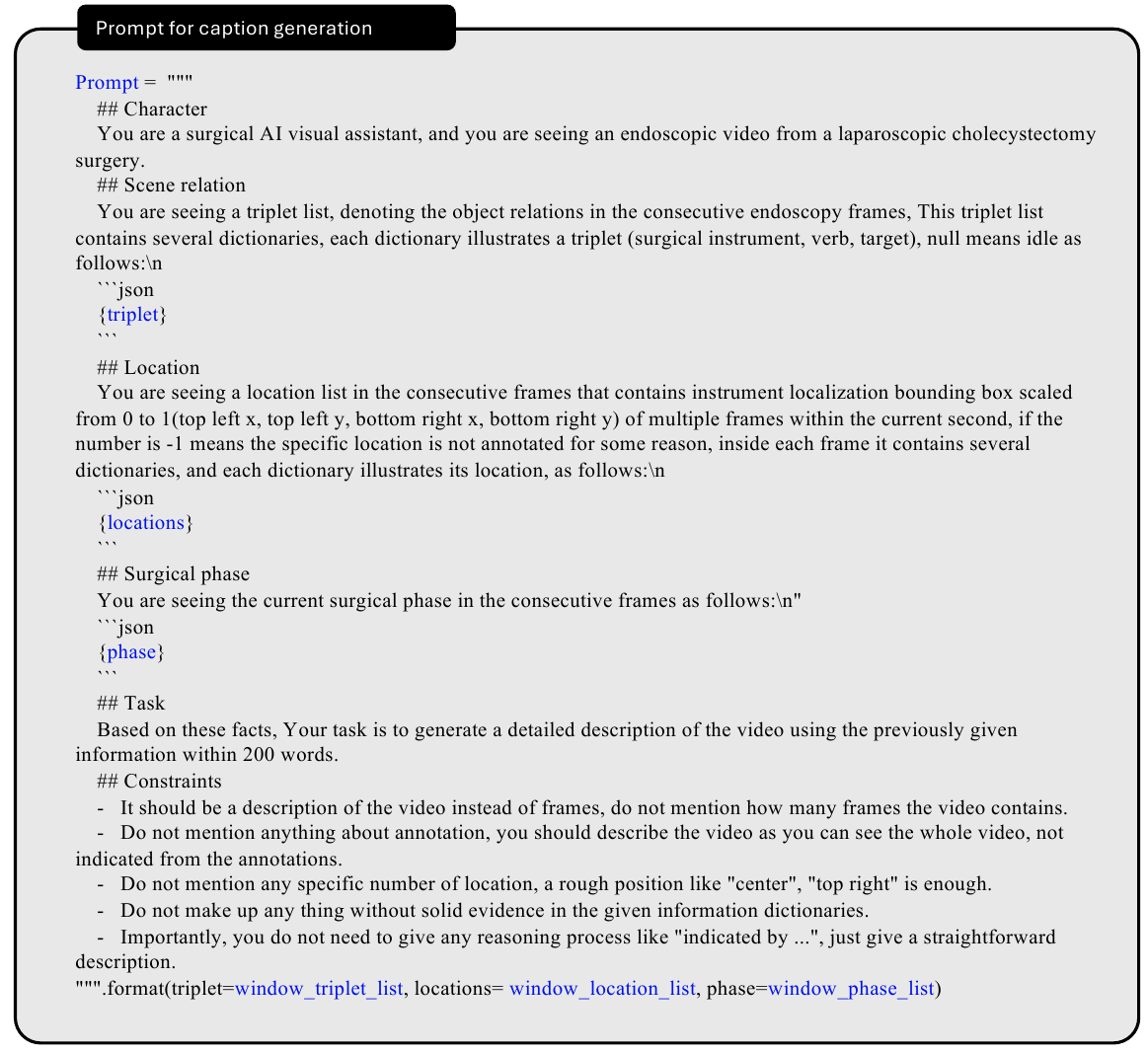}
    \caption{{The prompt for caption generation in our processed CholecT50 dataset.}}
    \label{fig_s10}
\end{figure*}

\begin{figure*}[h]
    \centering
    \includegraphics[width=1\textwidth]{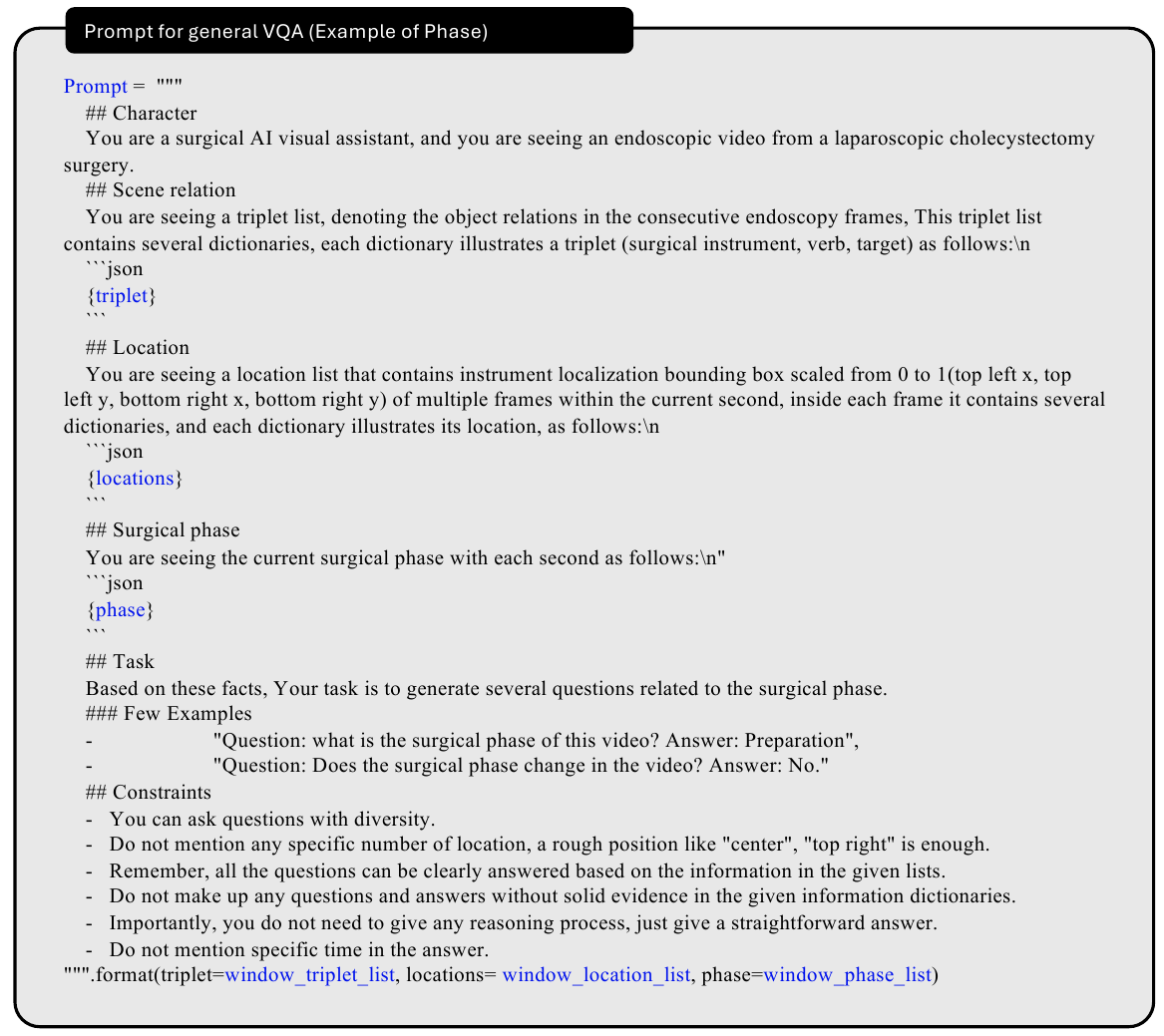}
    \caption{{The prompt for general VQA regarding the phase task in our processed CholecT50 dataset.}}
    \label{fig_s11}
\end{figure*}

\begin{figure*}[h]
    \centering
    \includegraphics[width=1\textwidth]{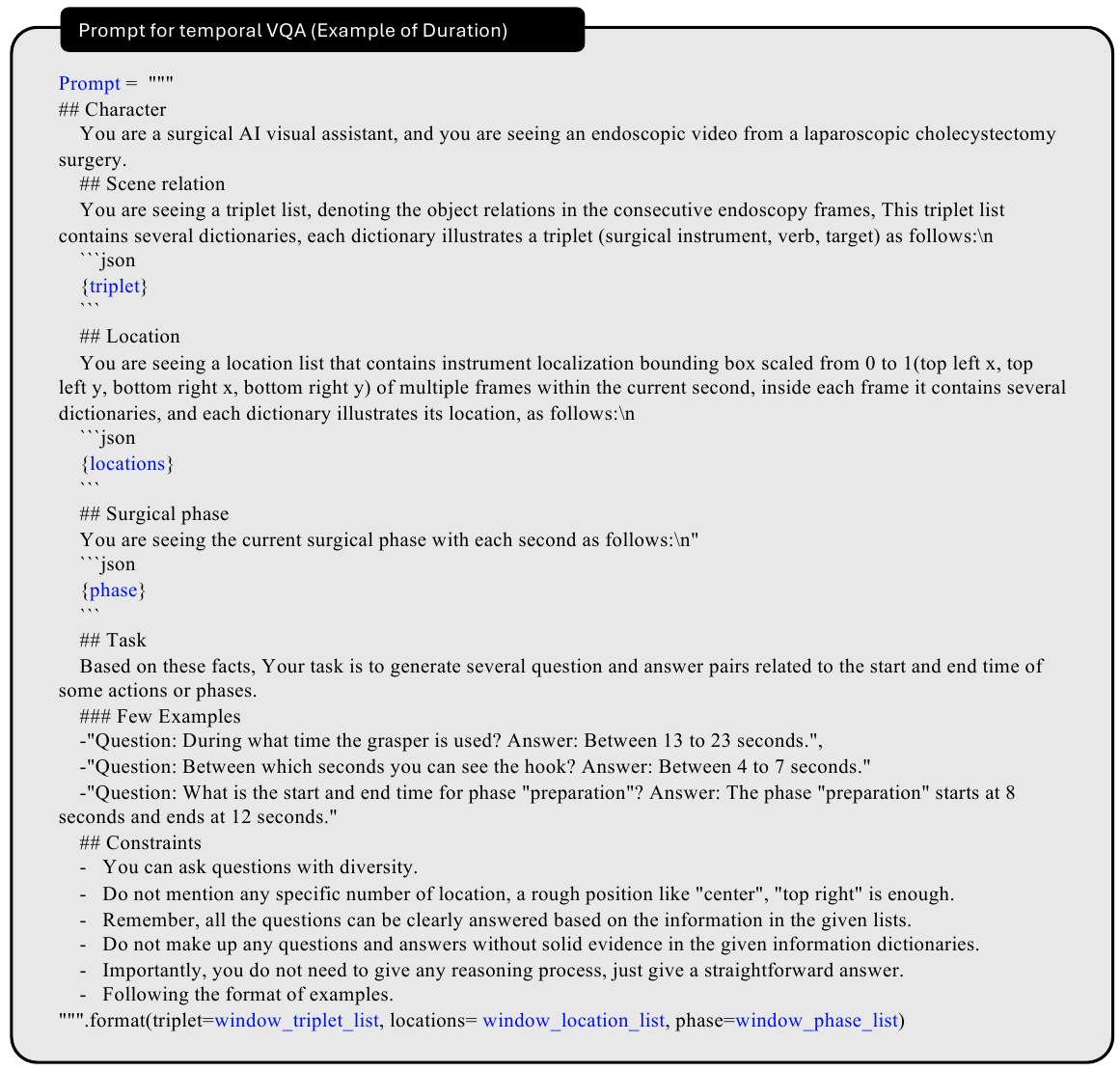}
    \caption{{The prompt for temporal VQA regarding the duration task in our processed CholecT50 dataset.}}
    \label{fig_s12}
\end{figure*}

\begin{figure*}[h]
    \centering
    \includegraphics[width=1\textwidth]{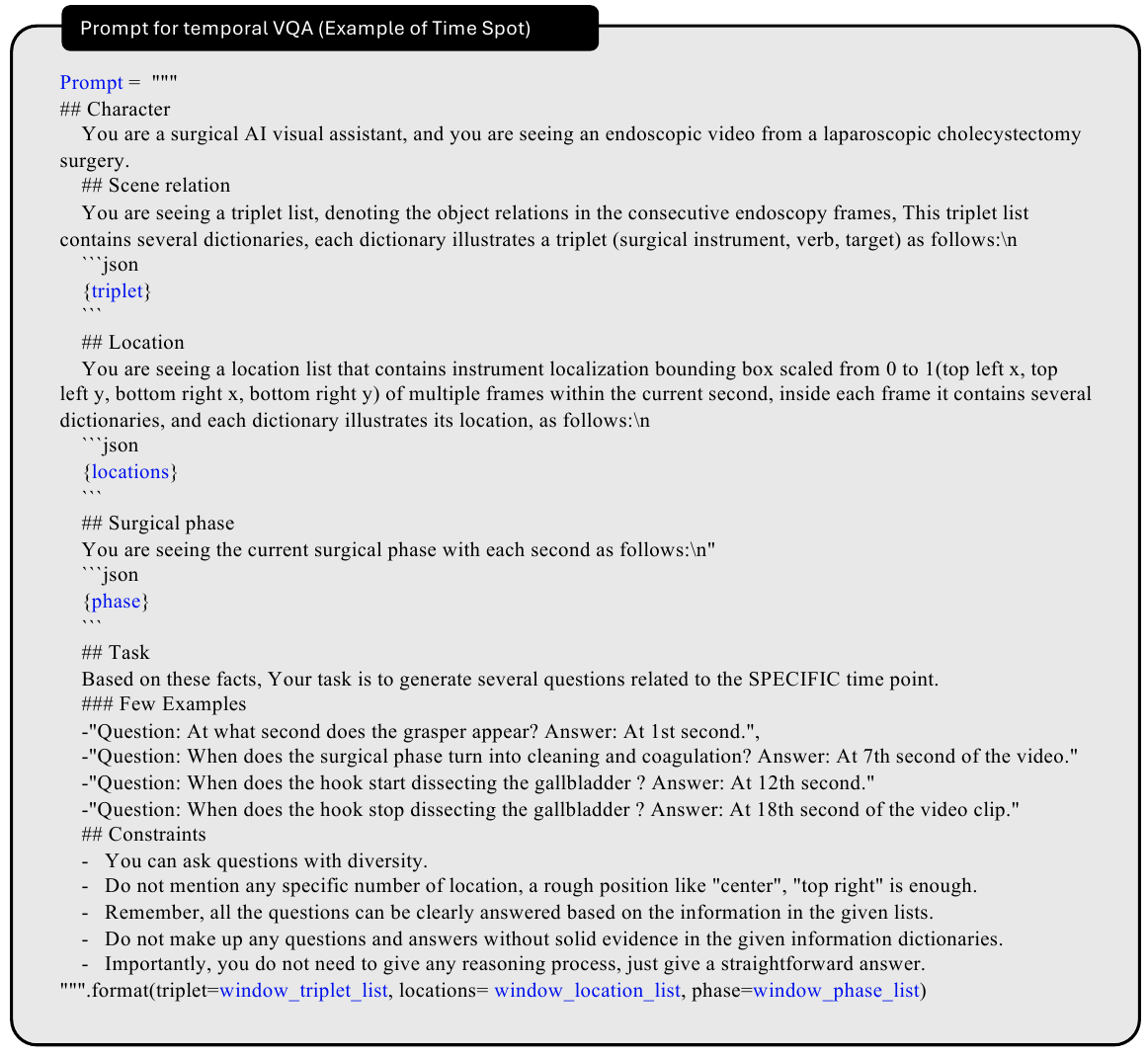}
    \caption{{The prompt for temporal VQA regarding the time spot task in our processed CholecT50 dataset.}}
    \label{fig_s13}
\end{figure*}

\end{document}